\ifdefined\pdfminorversion \pdfminorversion=7 \fi
\documentclass{article}

\PassOptionsToPackage{numbers, compress}{natbib}
\usepackage{ailab2026}

\usepackage[T1]{fontenc}
\usepackage{microtype}
\usepackage{url}
\usepackage{xurl}        % line breaks inside URLs
\usepackage{xspace}
\usepackage{soul}        % \hl for the highlighted key claim

\usepackage{amsmath}
\usepackage{amsfonts}
\usepackage{amssymb}
\usepackage{amsthm}

\usepackage{graphicx}
\usepackage{booktabs}
\usepackage{algorithm}
\usepackage{algorithmic}

\usepackage{caption}
\usepackage[hyperfootnotes=false]{hyperref}  % the author footnote has no mark to link
\hypersetup{colorlinks=true, citecolor=ailabgreen, linkcolor=ailabgreen, urlcolor=ailabgreen}

\colorlet{ailabgreenbgp}{ailabgreen!8}

\renewcommand{\oursrow}{\rowcolor{ailabgreenbgp}}
\newcommand{\bestm}[1]{\textcolor{ailabgreen}{\mathbf{#1}}}

\colorlet{ailabgreenhl}{ailabgreenbgp}
\renewcommand{\ailabemph}[1]{{\sethlcolor{ailabgreenhl}\hl{#1}}}
\newcommand{\ailabalgline}[1]{\makebox[0pt][l]{\setlength{\fboxsep}{1pt}\hspace*{-\fboxsep}\colorbox{ailabgreenbgp}{\makebox[\linewidth][l]{#1}}}}

\newcommand{\eg}{\textit{e.g.,}\xspace}

\newtheorem{lemma}{Lemma}

\newcommand{\keyfinding}[1]{}

\title{Factored One-sided Adam-Moment\\for Practical and Scalable SOAP}

\author{%
  Jaemyung Yu$^\star$\quad Byeongho Heo\quad Sangdoo Yun\quad Dongyoon Han$^\star$\\
  \\
  NAVER AI Lab
}

\hypersetup{
  pdftitle={Factored One-sided Adam-Moment for Practical and Scalable SOAP},
  pdfauthor={Jaemyung Yu, Byeongho Heo, Sangdoo Yun, Dongyoon Han},
}

\begin{document}

\maketitle
{\renewcommand{\thefootnote}{$\star$}\footnotetext{Correspondence to \{jaemyung.yu, dongyoon.han\}@navercorp.com}}

\begin{abstract}
Second-order optimizers (\eg Shampoo, SOAP) have emerged as compelling alternatives to Adam/AdamW and promise faster per-step convergence than AdamW in transformer training. However, they still remain costly in practice, and even SOAP still pays substantial optimizer-state memory and eigendecomposition overhead. We ask whether the benefits of SOAP can be retained at nearly the efficiency of AdamW. In this paper, we answer yes with a lightweight SOAP variant, Factored One-sided Adam-Moment (FOAM). The path to practical impact is twofold: first, FOAM employs input-only preconditioning, dropping the output-side Kronecker factor to reduce memory and remove one eigendecomposition. The Kronecker structure of the Hessian shows that this is exact for a diagonal output-side curvature and nearly lossless for a near-diagonal one, as we observe empirically. Second, FOAM replaces SOAP's dense rotated second moment with a factored Adam-moment estimator based on row-by-column statistics. Under the K-FAC Fisher assumption with a diagonal output curvature, this estimator exactly recovers SOAP's dense second moment, and we measure only a 15--20\% K-FAC violation, with negligible effect on the update. Our empirical results support the claim that FOAM matches strong SOAP and Muon baselines in validation loss across scales of large language model (LLM) pre-training, uses half of SOAP's optimizer state, runs at Muon-class wall-clock time, and retains robustness across a three-fold learning-rate range. The results hold from 124M to 760M parameters, demonstrating that SOAP's gains need not require SOAP's cost. Code: \url{https://github.com/naver-ai/foam}.

\end{abstract}

\section{Introduction}
\label{sec:intro}

{Adam~\citep{kingma2015adam} and AdamW~\citep{loshchilov2019adamw} have long been the default for pre-training transformer architectures, including large language models (LLMs). Their successors, including memory-efficient adaptive optimizers~\citep{zhang2025adammini,zhao2024galore,zhu2025apollo,gong2025alice}, show that comparable optimization quality can be achieved with substantially less optimizer memory. In another direction, Shampoo~\citep{gupta2018shampoo}, a foundational method in the family of Kronecker-factored second-order optimizers, has shown that the optimization quality of adaptive first-order methods such as Adam can be surpassed. Its most practical variant, SOAP~\citep{vyas2025soap}, the Adam-in-eigenbasis version of Shampoo, reports 35--40\% improvements in wall-clock time and step count over AdamW at large batch sizes, building upon the Distributed Shampoo baselines~\citep{anil2020scalable,shi2023distributed} that won the external-tuning track of the first AlgoPerf competition~\citep{dahl2023benchmarking, kasimbeg2025algoperf}.}

\begin{figure}[t]
\centering
\includegraphics{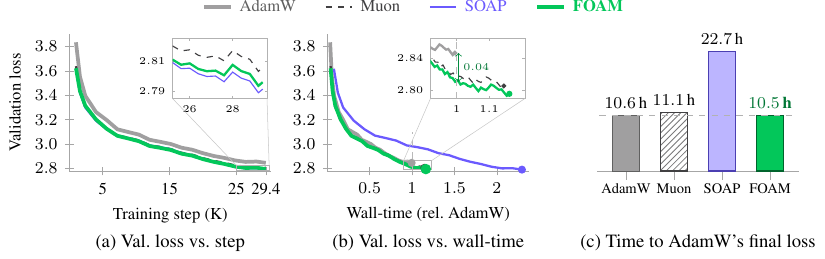}
\caption{\textbf{FOAM is a Pareto-efficient second-order optimizer.} GPT-2 large (760M), $29.55$K steps, each optimizer at its tuned learning rate. Curves are $3$-seed means smoothed over $1$K steps, and grey boxes mark the regions magnified in the insets. \textbf{(a) Per-step.} Validation loss against training step. In the last $4$K steps (inset) SOAP ends slightly below FOAM ($2.791$ vs.\ $2.796$, within about one seed std), both below Muon, while AdamW stays above the inset range. \textbf{(b) Wall-clock.} Same runs vs.\ wall time, normalized so AdamW finishes at $1.0\times$ (Muon, FOAM, and SOAP finish at $1.15\times$, $1.17\times$, and $2.31\times$). Within AdamW's own budget, FOAM is already $0.04$ nats below AdamW's final loss (inset), and it reaches SOAP's quality band at $51\%$ of SOAP's wall-clock time. \textbf{(c) Time to AdamW's final loss} ($2.847$; Table~\ref{tab:main}). FOAM gets there as fast as AdamW itself ($10.53$ vs.\ $10.55$\,h) and slightly ahead of Muon, while bilateral SOAP needs $2.15\times$ as long.}
\label{fig:teaser}
\vspace{-1em}
\end{figure}

{Still, practitioners often default to AdamW or Muon~\citep{jordan2024muon}, because SOAP achieves its gains at the expense of computational efficiency, requiring about $4.5{\times}$ the optimizer state of AdamW and a significantly slower per-step wall-clock time. More specifically, SOAP's memory is dominated by the first moment, the second moment in the eigenbasis-rotated space, and the two Kronecker factor matrices $L = G G^\top$ ($m^2$) and $R = G^\top G$ ($n^2$) of a gradient $G \in \mathbb{R}^{m \times n}$. Its runtime is further occupied by the periodic eigendecomposition of these factor matrices, which makes SOAP substantially more expensive than AdamW despite its strong optimization performance.}

{This paper aims to retain SOAP-level quality while approaching AdamW-level cost. Our motivation is simple. If SOAP's computational overhead can be sufficiently reduced, its optimization advantages could make it a practical successor to AdamW. To this end, we revisit SOAP's major cost components and show that several can be relaxed or replaced without sacrificing performance. The original SOAP paper~\citep{vyas2025soap} already evaluated one-sided and factored variants, but as cost-reducing ablations without a criterion for when they preserve quality. \citet{matrixwhitening2025} then found that input-only rotation and a factored variance each retain most of SOAP's quality, and other works~\citep{lin2025kl, 2025purifying} improved SOAP along other axes. Our contribution is thus not one-sided preconditioning or factorization itself, but explaining \emph{which side to retain}, \emph{when factorization is justified}, and \emph{why their combination is effective}.}

Concretely, our approach is built on two structural simplifications: input-only preconditioning and a factored Adam-Moment update. The first simplification drops the output-side factor $L$ and retains only the input-side factor $R$, eliminating its $m^2$ memory and one of the two eigendecompositions. Its justification follows from the Kronecker structure of the Hessian, in which all output rows of an affine layer share the same input-side outer product $x x^\top$~\citep{kunstner2024whyadam}, while the Hessian of transformer layers is empirically near-block-diagonal along output neurons or heads~\citep{zhang2025adammini}, making the input-only form the natural one-sided approximation. The second simplification replaces the dense rotated second moment with a row-and-column factored estimator, $r \in \mathbb{R}^m$ and $c \in \mathbb{R}^n$, with trace normalization, reducing the buffer size from $mn$ to $m+n$. Under the K-FAC Fisher assumption~\citep{martens2015kfac} with a diagonal output curvature, this estimator exactly recovers SOAP's dense second moment, and the K-FAC part holds to a $15$--$20\%$ relative error in practice. Together, these simplifications reduce SOAP's optimizer state from $2mn + 2m^2 + 2n^2$ to $mn + 2n^2 + m + n$ and the amortized eigendecomposition count from two to one. We call the resulting method \emph{FOAM} (\textbf{F}actored \textbf{O}ne-sided \textbf{A}dam-\textbf{M}oment).

Figure~\ref{fig:teaser} highlights FOAM's main advantages. On pre-training, \ailabemph{FOAM matches the strong Muon and SOAP baselines on validation loss with half of SOAP's optimizer state and reaches AdamW's final loss within $7\%$ of AdamW's own wall-clock time}, while SOAP needs more than twice as long at 760M, and FOAM matches AdamW's hyperparameter robustness over a three-fold learning-rate range. The advantage holds from \texttt{124M} to \texttt{760M} parameters with learning rates re-tuned at every size from a $\mu$P~\citep{yang2022tensor} anchor. In summary, our contributions are: (i) a structural justification of input-only preconditioning, with the Kronecker structure of the Hessian and empirical verifications with LLMs with recent architectural elements; (ii) a factored Adam-Moment update with trace normalization, exact under K-FAC with a diagonal output curvature and empirically validated through K-FAC-residual measurements during training; and (iii) the resulting algorithm FOAM, a scalable and practical SOAP variant that delivers SOAP-class quality with half of SOAP's optimizer state ($+7.4\%$ measured peak training memory over AdamW) and Muon-class wall-clock time.

\section{Related Work}
\label{sec:related}

Adaptive optimizers can be categorized by how aggressively they approximate the empirical Fisher $F = \mathbb{E}[\mathrm{vec}(G)\,\mathrm{vec}(G)^\top]$, which all of Adam~\cite{kingma2015adam}/AdamW~\cite{loshchilov2019adamw}, Shampoo~\cite{gupta2018shampoo}, and SOAP~\cite{vyas2025soap} implicitly use as a curvature surrogate in place of the true Fisher~\citep{kunstner2019empirical}.

\subsection{Empirical Fisher and Kronecker-factored Optimizers}
\label{sec:related:kfac}

AdaGrad~\citep{duchi2011adaptive} introduced per-coordinate scaling by an empirical second-moment statistic, an early diagonal approximation of the inverse empirical Fisher. K-FAC~\citep{martens2015kfac} replaces the diagonal with a Kronecker product, $F \approx \mathbb{E}[a a^\top] \otimes \mathbb{E}[s s^\top]$ where $a$ is the layer input and $s$ the gradient with respect to the layer's pre-activation output, the assumption under which our factored Adam-Moment update is exact. \citet{eschenhagen2023kfac} adapt K-FAC to transformer architectures via the \emph{expand} and \emph{reduce} flavors. Shampoo~\citep{gupta2018shampoo} carries this idea to the matrix parameters. It maintains EMAs of $L = G G^\top$ and $R = G^\top G$ and steps in $L^{-1/4} G R^{-1/4}$, i.e., it preconditions by $(R \otimes L)^{-1/4}$; \citet{morwani2024shampoo} show that $R \otimes L$ equals one power-iteration step toward the optimal Kronecker-product approximation of the empirical Fisher and is empirically close to it. SOAP~\citep{vyas2025soap} eigendecomposes $L$ and $R$ and runs Adam in the resulting rotated coordinates, an EKFAC-style construction~\citep{george2018ekfac} that~\citet{lu2025soap} interpret as gradient whitening. Variants modify specific pieces: CASPR~\citep{duvvuri2024caspr} combines the per-axis factors through a Kronecker sum instead of a Kronecker product. KL-Shampoo and KL-SOAP~\citep{lin2025kl} recast the family via KL-divergence minimization, and Purifying Shampoo~\citep{2025purifying} shows that trace scaling of the preconditioner can replace learning-rate grafting, which corroborates our trace-norm choice in \S\ref{sec:method:c2}. The one-sided fallback in distributed Shampoo~\citep{anil2020scalable} drops a factor on memory grounds, whereas we give a structural argument for keeping the input side (\S\ref{sec:method:c1}). Purifying Shampoo's adaptive tolerance criterion for skipping the periodic eigendecomposition is adopted unchanged as FOAM's amortized eigenbasis-refresh policy (\S\ref{sec:method:c2}).

\subsection{Optimizers with Coarser Approximations}
\label{sec:related:approx}

A second line of work adopts coarser approximations to the empirical Fisher in exchange for lower computational cost. Adam~\citep{kingma2015adam,loshchilov2019adamw} keeps only the diagonal of $F$, sacrificing all matrix structure. Signum~\citep{bernstein2018signum} keeps only the sign of the momentum. Muon~\citep{jordan2024muon} replaces the second-moment buffer with Newton-Schulz orthogonalization $M (M^\top M)^{-1/2}$, which can be read as a square-root natural-gradient step under a structured Fisher $\tilde R \otimes I_m$ with an input-side factor $\tilde R \in \mathbb{R}^{n \times n}$ shared across output rows~\citep{ma2024swan, gong2025alice}. AdaMuon~\citep{si2025adamuon} adds an element-wise Adam-style second moment on top of the orthogonalized update. SPlus~\citep{frans2025splus}, a related SOAP variant, replaces the rotated second moment with sign normalization in the Shampoo eigenbasis and adds iterate averaging. Recent Muon variants (NorMuon~\citep{li2026normuon}, Muon$^2$~\citep{liu2026muon2}, NAMO~\citep{zhang2026namo}, Polar Express~\citep{amsel2026polar}) refine specific components of Muon. A parallel branch targets memory-efficient AdamW rather than matching SOAP-class quality: Adam-mini~\citep{zhang2025adammini}, GaLore~\citep{zhao2024galore}, APOLLO~\citep{zhu2025apollo}, Alice~\citep{gong2025alice}, and SWAN~\citep{ma2024swan}. We exclude these methods from the main comparison since they mainly target a smaller memory footprint than AdamW, whereas we target SOAP-level performance at near-Adam cost.

\noindent\textbf{Positioning our method.} FOAM lies at the intersection of these two branches. It stays within the Kronecker-factored Fisher framework but borrows the factored-second-moment idea from Adafactor~\citep{shazeer2018adafactor}'s row-by-column estimator, applied here in the rotated coordinates of SOAP. Both ingredients appear in the SOAP paper as cost-reducing ablations, a one-sided variant and an Adafactor-style factored second moment in the eigenbasis, and \citet{matrixwhitening2025} found empirically that input-only rotation and a factored variance each preserve most of SOAP's quality; the distinction lies in the criteria. SOAP's one-sided variant keeps the side with the smaller dimension, whereas we keep the input side irrespective of size (\S\ref{sec:method:c1}), so the two rules choose opposite sides for the MLP down-projection; likewise, we factorize only after the input-side rotation, which makes the rotated second moment separable into row and column factors (Lemma~\ref{lem:rotcov}, \S\ref{sec:method:c2}).

\section{Preliminary: Shampoo and SOAP}
\label{sec:preliminary}
We consider a transformer with matrix parameters $W \in \mathbb{R}^{m \times n}$, where $n$ and $m$ denote the input and output dimensions, respectively. Let $G_t \in \mathbb{R}^{m \times n}$ denote the mini-batch gradient at step $t$. Following the convention of SOAP~\citep{vyas2025soap}, we define the two Kronecker factors $L_t$ and $R_t$ as the second-moment matrices of the gradient:
\begin{equation}
L_t \;=\; \mathbb{E}[G_t G_t^\top] \;\in\; \mathbb{R}^{m \times m}, \qquad
R_t \;=\; \mathbb{E}[G_t^\top G_t] \;\in\; \mathbb{R}^{n \times n},
\end{equation}
$L_t$ acts on the left side (\textbf{output}) of $G_t$, and $R_t$ acts on the right side (\textbf{input}). In practice, both are estimated by exponential moving averages (EMAs), for which we reuse the same symbols, and we reserve the term \emph{estimator} for such EMA buffers. Shampoo~\citep{gupta2018shampoo} takes a fractional inverse of each factor and applies the preconditioned update:
\begin{figure}[t]
\centering
\includegraphics{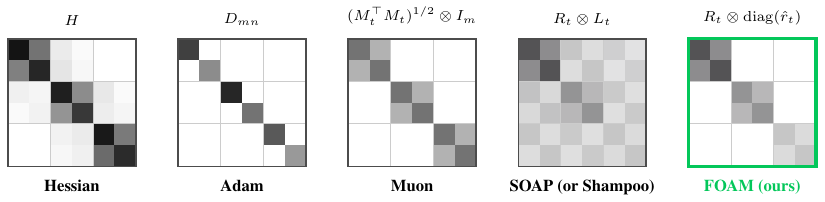}
\caption{\textbf{Hessian and its approximation across optimizers.} For $W \in \mathbb{R}^{m \times n}$ with $m{=}3, n{=}2$, we visualize the $mn \times mn$ Hessian alongside the preconditioner of each optimizer; shaded cells are non-zero and darker means larger; coordinates are grouped by output row ($n \times n$ blocks), while labels follow the column-major convention of Lemma~\ref{lem:rotcov}. The Hessian is near-block-diagonal. \textbf{Adam} ($D_{mn}$): fully diagonal approximation with no Kronecker structure. \textbf{Muon} ($(M_t^\top M_t)^{1/2} \otimes I_m$): $m$ identical outer-diagonal blocks, built statelessly from the current momentum via $M_t (M_t^\top M_t)^{-1/2}$. \textbf{SOAP} ($R_t \otimes L_t$): full Kronecker fit with EMAs on both sides, with all $m^2$ outer blocks populated. \textbf{FOAM} ($R_t \otimes \mathrm{diag}(\hat r_t)$, ours): replaces Muon's identity output with a per-block diagonal scaling, and its stateless input factor with the EMA of $G_t^\top G_t$.}
\label{fig:concept}
\end{figure}

\begin{equation}
W_{t+1} \;=\; W_t \;-\; \eta \, L_t^{-1/4} \, G_t \, R_t^{-1/4},
\end{equation}
with learning rate $\eta$. SOAP~\citep{vyas2025soap} instead eigendecomposes $L_t$ and $R_t$ to obtain orthogonal rotations $Q_L \in \mathbb{R}^{m \times m}$ and $Q_R \in \mathbb{R}^{n \times n}$, and runs Adam in the rotated coordinates:
\begin{equation}
G_{\rm rot} \;=\; Q_L^\top \, G \, Q_R \;\in\; \mathbb{R}^{m \times n},
\end{equation}
following the EKFAC tradition~\citep{george2018ekfac} that tracks diagonal second moments in the eigenbasis of a Kronecker-factored preconditioner. The per-step update maintains a dense second moment $v_{\rm rot} \in \mathbb{R}^{m \times n}$ in the rotated space. The rotations $Q_L$ and $Q_R$ are refreshed every $\mathtt{pf}$ (precondition frequency) steps via two eigendecompositions of $L_t$ and $R_t$. $L_t$ and $R_t$ require $m^2 + n^2$ memory, the rotated second moment $v_{\rm rot}$ requires an additional $mn$ memory, and the periodic eigendecompositions incur further computational overhead. These costs constitute the overhead structure that FOAM addresses.

\section{Factored One-sided Adam-Moment (FOAM)}
\label{sec:method}

We propose \textbf{FOAM}, \textbf{F}actored \textbf{O}ne-sided \textbf{A}dam-\textbf{M}oment, which combines two structural simplifications of SOAP into a single optimizer. FOAM applies an input-side K-FAC factor combined with a factored output diagonal. It maintains only the input-side factor $R_t$, runs Adam in the eigenbasis $Q_R = \mathrm{eigh}(R_t)$, and replaces SOAP's dense rotated second moment $v_{\rm rot} \in \mathbb{R}^{m \times n}$ with a row-column factored estimator $\hat v_{\rm rot} = \tau \, \hat r \, \hat c^{\top}$. Here $\hat r \in \mathbb{R}^m$ and $\hat c \in \mathbb{R}^n$ are bias-corrected EMAs of the row and column means of the squared rotated gradient, and $\tau$ is a trace-matching scalar. The two simplifications together give FOAM the implicit Fisher $R_t \otimes \mathrm{diag}(\hat r_t)$, a full input-side factor with a diagonal output side, derived in Algorithm~\ref{alg:foam} and the following sections. Figure~\ref{fig:concept} compares this preconditioner structure with those of Adam, Muon, and SOAP against the near-block-diagonal Hessian of a transformer layer, where FOAM keeps SOAP's input factor with a diagonal output side.

\noindent\textbf{Derivation overview.}
The steps below form a single chain from $\mathbb{E}[gg^\top]$, $g = \mathrm{vec}(G)$, to the two stored vectors.
(i) Under K-FAC, $\mathbb{E}[gg^\top] = R \otimes L$, and $L$ is near-diagonal (\S\ref{sec:method:c1}), so the off-diagonal structure lives on the input side, which is the side to keep.
(ii) Rotating by $Q_R$ turns the covariance into $\Lambda_R \otimes L$ (Lemma~\ref{lem:rotcov}). Like SOAP running Adam in its eigenbasis, FOAM keeps only the diagonal of this covariance, a modeling choice that the rotation makes accurate.
(iii) This diagonal, reshaped to $m \times n$, is the rank-one matrix $\mathrm{diag}(L)\, \lambda_R^\top$, so storing $\hat r$ and $\hat c$ keeps every coordinate's own variance and loses nothing relative to the dense $mn$ buffer, as formalized in \S\ref{sec:method:c2}.

\begin{algorithm}[t]
\caption{\textbf{\textsc{FOAM} update at step $t$ for $W \in \mathbb{R}^{m \times n}$.} $\mathbf{1}_d$ is the all-ones vector in $\mathbb{R}^d$, and $G^{\circ 2}$, $\oslash$, and $\sqrt{\cdot}$ act element-wise. $\mathrm{off}(\cdot)$ is the off-diagonal ratio of a matrix (\S\ref{sec:method:c2}). $Q_R$ is initialized to the identity, and decoupled weight decay is applied as in AdamW and omitted here.}
\label{alg:foam}
\begin{algorithmic}[1]
\REQUIRE gradient $G$; hyperparameters $(\eta, \beta_1, \beta_2, \varepsilon, \mathtt{pf}, \mathrm{tol})$; state $M, r, c, R, Q_R$
\STATE $R \gets \beta_2 R + (1 - \beta_2)\, G^\top G$
\hfill\textit{(input-side factor)}
\IF{$t \bmod \mathtt{pf} = 0$ \textbf{and} $\mathrm{off}(Q_R^\top R\, Q_R) > \mathrm{tol}$}
\STATE $Q_R \gets \mathrm{eigh}(R)$
\hfill\textit{(refresh the eigenbasis)}
\ENDIF
\STATE $M \gets \beta_1 M + (1 - \beta_1)\, G$
\hfill\textit{(first moment)}
\STATE \ailabalgline{$G_{\rm rot} \gets G\, Q_R$,\; $M_{\rm rot} \gets M\, Q_R$
\hfill\textit{(rotate into the eigenbasis)}}
\STATE $r \gets \beta_2 r + (1 - \beta_2)\, G_{\rm rot}^{\circ 2}\, \mathbf{1}_n / n$,\;
$c \gets \beta_2 c + (1 - \beta_2)\, \mathbf{1}_m^\top G_{\rm rot}^{\circ 2} / m$
\hfill\textit{(factored second moment)}
\STATE $\hat M_{\rm rot} \gets M_{\rm rot} / (1 - \beta_1^t)$,\;
$\hat r \gets r / (1 - \beta_2^t)$,\;
$\hat c \gets c / (1 - \beta_2^t)$
\hfill\textit{(bias correction)}
\STATE \ailabalgline{$\hat v_{\rm rot} \gets \tau\, \hat r\, \hat c^{\top}$ with $\tau = \mathrm{tr}(R) / (\|\hat r\|_1 \|\hat c\|_1)$
\hfill\textit{(rank-one second moment at $R$'s scale)}}
\STATE $U_{\rm rot} \gets \hat M_{\rm rot} \oslash (\sqrt{\hat v_{\rm rot}} + \varepsilon)$
\hfill\textit{(Adam step in the eigenbasis)}
\STATE $W \gets W - \eta\, U_{\rm rot}\, Q_R^\top$
\hfill\textit{(rotate back and update)}
\end{algorithmic}
\end{algorithm}

\subsection{Input-only Preconditioning}
\label{sec:method:c1}

Our idea is to remove the output-side factor $L_t$ and maintain only $R_t \leftarrow \beta_2 R_{t-1} + (1 - \beta_2)\, G_t^\top G_t$ with its eigenbasis $Q_R = \mathrm{eigh}(R_t)$. The gradient rotation is applied only to the input axis, i.e., $G_{\rm rot} = G\, Q_R \in \mathbb{R}^{m \times n}$. This reduces the factor-state memory from $\mathcal{O}(m^2 + n^2)$ to $\mathcal{O}(n^2)$ and the number of eigendecompositions from two to one, halving this cost for square matrices.

A conceptual justification comes from the curvature of an affine layer with cross-entropy loss, the standard loss for language modeling and classification. For $y = Wx$ and $p = \mathrm{softmax}(y)$, the Hessian of the loss $\ell = -\log p_{k^\star}$ for a target class $k^\star$ with respect to $W$ decomposes exactly as
\begin{equation}
H \,=\, xx^\top \otimes \big(\mathrm{diag}(p) - pp^\top\big).
\end{equation}
The output-side factor $\mathrm{diag}(p) - pp^\top$ has diagonal entries $p_i(1 - p_i)$ and off-diagonal entries $-p_i p_k$. Each off-diagonal entry $p_i p_k$ is smaller than the diagonal entry $p_i(1 - p_i)$ by a factor of $p_k/(1 - p_i)$, which is small for most classes of a large vocabulary even far from a one-hot prediction, in line with the dominant per-class diagonal blocks of softmax linear models~\citep{kunstner2024whyadam}. Larger entries arise only where a class takes much of the mass, and each pair $(i,k)$ receives them only from examples in which $i$ or $k$ is such a class, whereas the diagonal accumulates over all examples, so with a large vocabulary the expected output-side curvature is dominated by its diagonal, and under the K-FAC independence approximation the batch expectation becomes
\begin{equation}
\mathbb{E}[H] \;\approx\; \mathbb{E}[xx^\top] \otimes \mathrm{diag}(\kappa), \qquad \kappa_i \;=\; \mathbb{E}\big[p_i(1 - p_i)\big].
\end{equation}
This has the structure of FOAM's implicit Fisher, with the input-side curvature $\mathbb{E}[xx^\top]$ tracked by $R_t$ and the per-output curvature scale $\kappa_i$ absorbed into $\hat r_t$. The output side need not be homogeneous across rows. FOAM only requires that it is dominated by its diagonal, which this mechanism suggests and the Gauss-Newton measurements of \S\ref{sec:exp:analysis} support empirically.

\noindent\textbf{Beyond the affine setting.} For a hidden affine layer $h = Wx$, the per-example Hessian keeps the same form $xx^\top \otimes \nabla_h^2 \ell$, since $h$ is linear in $W$, and only the output-side factor changes, from $\mathrm{diag}(p) - pp^\top$ to the curvature with respect to $h$. The vanishing-off-diagonal mechanism is specific to the output layer, but near-block-diagonal Hessians have been observed in MLPs throughout training~\citep{collobert2004large, zhang2025adammini} and, with blocks along output neurons or heads, in a small transformer~\citep{zhang2025adammini}. We provide additional indirect evidence specific to our setting through Gauss-Newton alignment measurements in \S\ref{sec:exp:analysis}.

\subsection{Factored Eigenvalue Under Input-only Preconditioning}
\label{sec:method:c2}
\begin{table}[t]
\centering
\caption{\textbf{Per-parameter complexity.} State, per-step compute, and amortized eigendecomposition cost for $W \in \mathbb{R}^{m \times n}$ with output-side $L \in \mathbb{R}^{m \times m}$ and input-side $R \in \mathbb{R}^{n \times n}$. FOAM keeps Muon's per-step cost.}
\label{tab:complexity}
\small
\begin{tabular}{lcccc}
\toprule
& \textbf{AdamW} & \textbf{Muon} & \textbf{SOAP} & \textbf{FOAM} \\
\midrule
Optimizer state & $2 m n$ & $m n$ & $2 m n + 2 m^2 + 2 n^2$ &
$m n + 2 n^2 + (m + n)$ \\
Per-step cost & $\mathcal{O}(m n)$ & $\mathcal{O}(m n^2)$ &
$\mathcal{O}(m n (m + n))$ & $\mathcal{O}(m n^2)$ \\
Amortized (per $\mathtt{pf}$) & -- & -- &
$\mathcal{O}(m^3 + n^3)$ & $\mathcal{O}(n^3)$ \\
\bottomrule
\end{tabular}
\end{table}

Even with the input-only simplification, SOAP still has overheads due to the dense rotated second-moment buffer $v_{\rm rot} \in \mathbb{R}^{m \times n}$. Our next step is to replace it with two factored estimators $r \in \mathbb{R}^m$ and $c \in \mathbb{R}^n$, EMAs of the row- and column-means of $G_{\rm rot}^{\circ 2} = G_{\rm rot} \circ G_{\rm rot}$, and reconstruct the rotated second moment as an outer product with trace normalization:
\begin{equation}
\hat v_{\rm rot}[i, j] \;=\; \tau \cdot \hat r[i] \cdot \hat c[j], \qquad
\tau \;=\; \frac{\mathrm{tr}(R)}{\|\hat r\|_1\, \|\hat c\|_1}.
\end{equation}
The state shrinks from $mn$ to $m + n$ per matrix parameter. The trace-matching scalar $\tau$ makes the entries of $\hat v_{\rm rot}$ sum to $\mathrm{tr}(R)$, fixing the overall scale of the reconstructed second moment to that of the input-side factor. We now justify why this rank-one factorization is sufficient.

\begin{lemma}[Covariance of the input-rotated gradient]
\label{lem:rotcov}
Let $g = \mathrm{vec}(G) \in \mathbb{R}^{mn}$ and assume the K-FAC factorization $\mathbb{E}[gg^\top] = R \otimes L$. Let $g' = \mathrm{vec}(G\, Q_R)$ be the input-rotated gradient in column-major form. Then
\begin{equation}
\mathbb{E}[g'\, g'^\top] \;=\; \Lambda_R \otimes L,
\end{equation}
where $\Lambda_R = Q_R^\top R\, Q_R$ is the diagonal eigenvalue matrix of $R$.
\end{lemma}

\begin{proof}
Using the column-major identity $\mathrm{vec}(AGB) = (B^\top \otimes A)\, \mathrm{vec}(G)$ with $A = I_m$ and $B = Q_R$, we have $g' = (Q_R^\top \otimes I_m)\, g$. Therefore,
\begin{equation}
\mathbb{E}[g'\, g'^\top] \;=\; (Q_R^\top \otimes I_m)\, \mathbb{E}[g\, g^\top]\, (Q_R \otimes I_m) \;=\; (Q_R^\top \otimes I_m)(R \otimes L)(Q_R \otimes I_m).
\end{equation}
The Kronecker mixed-product identity $(A \otimes B)(C \otimes D) = (AC) \otimes (BD)$, applied twice, gives
\begin{equation}
(Q_R^\top \otimes I_m)(R \otimes L)(Q_R \otimes I_m) \;=\; (Q_R^\top R\, Q_R) \otimes (I_m\, L\, I_m) \;=\; \Lambda_R \otimes L. \qedhere
\end{equation}
\end{proof}

Lemma~\ref{lem:rotcov} tells us two things about the rotated gradient. First, the input-side covariance becomes diagonal, because rotating by $Q_R$ decorrelates the input directions and replaces $R$ with its eigenvalues $\Lambda_R$. Second, the output-side factor $L$ is unchanged, so when $L$ is diagonal (\S\ref{sec:method:c1}) the rotated covariance $\Lambda_R \otimes L$ is itself diagonal, and its diagonal, the part FOAM keeps, factorizes as
\begin{equation}
\mathbb{E}\big[(G\, Q_R)[i, j]^2\big] \;=\; \lambda_R[j] \cdot L[i, i],
\end{equation}
with $\lambda_R[j]$ the $j$-th diagonal entry of $\Lambda_R$. The right-hand side is a per-row factor depending only on $i$ times a per-column factor depending only on $j$, so the entire $m \times n$ second-moment surface is captured by $m + n$ numbers rather than $mn$. The row- and column-means of $G_{\rm rot}^{\circ 2}$ are the natural estimators of these two factors, up to a global scalar that the trace normalization $\tau$ removes, so $\tau\, \hat r\, \hat c^{\top}$ recovers the dense $mn$ buffer exactly under K-FAC, and SOAP's two-sided one when $L$ is also diagonal. In our runs, the K-FAC assumption is violated by a $15$--$20\%$ relative error, mostly at the key projection, yet this costs only about $0.002$ nats in final validation loss (Table~\ref{tab:e3-stack}, Appendix~\ref{sec:appx:kfac}).

\noindent\textbf{Amortized eigenbasis update.} The eigenbasis $Q_R$ is refreshed every $\mathtt{pf}$ steps via the adaptive skip rule of Purifying Shampoo~\citep{2025purifying}. We form $B = Q_R^\top R_t Q_R$ and skip the eigendecomposition whenever its off-diagonal ratio $\mathrm{off}(B) = \| B - \mathrm{diag}(B) \|_F / \| B \|_F \le \mathrm{tol}$, keeping the previous $Q_R$ ($\mathrm{tol} = 0$ refreshes at every $\mathtt{pf}$-th step). With a small $\mathrm{tol}$, an eigendecomposition is invoked only when $R_t$ has drifted from diagonal in the current basis, which becomes rare late in training.

\noindent\textbf{Complexity.}
\label{sec:method:complexity}
Table~\ref{tab:complexity} compares per-parameter optimizer state and compute. SOAP stores $2mn + 2m^2 + 2n^2$ values for the first and rotated second moments, two factor EMAs, and two eigenbases. FOAM reduces this to $mn + 2n^2 + (m + n)$ for one first moment, the estimators $r, c$, and the factor $R$ with its eigenbasis $Q_R$. Per-step cost is $\mathcal{O}(mn^2)$ from the rotation $G\, Q_R$, matching Muon, and the amortized eigendecomposition cost drops from $\mathcal{O}(m^3 + n^3)$ to $\mathcal{O}(n^3)$.

\section{Experiment}
\label{sec:exp}
This section asks whether FOAM matches strong Muon and SOAP baselines in pre-training across scales (\S\ref{sec:exp:main}), and whether its design components are well-founded and each contributes (\S\ref{sec:exp:analysis}).

\subsection{Experimental Setup}
\label{sec:exp:setup}

\begin{table}[t]
\centering
\caption{\textbf{Pre-training results.} For each method and scale at the best LR/WD (details in Appendix~\ref{sec:appx:protocol}) of its own search, we report validation loss (\emph{val}, mean $\pm$ std over $11$ seeds and $3$ at large), matrix-group optimizer state in GiB (\emph{state}), per-step wall time in ms on one $8{\times}$H100 node (\emph{iter}), and the wall-clock $t_{\mathrm{A}}$ (minutes at small, hours otherwise) to reach AdamW's per-size final loss, i.e., the steps to reach it times \emph{iter}.}
\label{tab:main}
\footnotesize
\tabcolsep=0.4em
\resizebox{\linewidth}{!}{%
\begin{tabular}{l cccc cccc cccc}
\toprule
& \multicolumn{4}{c}{\textbf{small} (124M)}
& \multicolumn{4}{c}{\textbf{medium} (350M)}
& \multicolumn{4}{c}{\textbf{large} (760M)} \\
\cmidrule(lr){2-5} \cmidrule(lr){6-9} \cmidrule(lr){10-13}
Optimizer
& \emph{val} & \emph{state} & \emph{iter} & $t_{\mathrm{A}}$
& \emph{val} & \emph{state} & \emph{iter} & $t_{\mathrm{A}}$
& \emph{val} & \emph{state} & \emph{iter} & $t_{\mathrm{A}}$ \\
\midrule
AdamW
& $3.370{\scriptstyle\pm.009}$ & $0.63$ & $311$ & $23.8$
& $3.040{\scriptstyle\pm.012}$ & $2.25$ & $856$ & $3.04$
& $2.847{\scriptstyle\pm.004}$ & $5.27$ & $1449$ & $10.55$ \\
Muon
& $3.313{\scriptstyle\pm.008}$ & $0.32$ & $361$ & $22.9$
& $2.998{\scriptstyle\pm.010}$ & $1.12$ & $983$ & $2.95$
& $2.806{\scriptstyle\pm.005}$ & $2.64$ & $1668$ & $11.12$ \\
AdaMuon
& $3.311{\scriptstyle\pm.009}$ & $0.63$ & $366$ & $23.2$
& $2.992{\scriptstyle\pm.012}$ & $2.25$ & $991$ & $2.97$
& $2.799{\scriptstyle\pm.004}$ & $5.27$ & $1684$ & $10.85$ \\
SPlus
& $3.325{\scriptstyle\pm.007}$ & $2.85$ & $525$ & $35.0$
& -- & -- & -- & --
& -- & -- & -- & -- \\
SOAP
& $\bestm{3.305}{\scriptstyle\pm.009}$ & $2.85$ & $536$ & $32.1$
& $\bestm{2.983}{\scriptstyle\pm.012}$ & $10.12$ & $1521$ & $4.56$
& $\bestm{2.791}{\scriptstyle\pm.005}$ & $23.73$ & $3343$ & $22.66$ \\
\oursrow
\textbf{FOAM (ours)}
& $3.309{\scriptstyle\pm.008}$ & $1.46$ & $393$ & $24.9$
& $2.991{\scriptstyle\pm.012}$ & $5.06$ & $1083$ & $3.25$
& $2.796{\scriptstyle\pm.005}$ & $11.87$ & $1692$ & $10.53$ \\
\bottomrule
\end{tabular}
}
\end{table}

\noindent\textbf{Setup.} We pre-train the models following a modern LLM architectural setup (RoPE~\citep{su2024roformer}, RMSNorm~\citep{zhang2019rmsnorm}, untied embeddings, MLP with GELU~\citep{hendrycks2016gelu}, no bias) at three scales following GPT-2~\citep{radford2019language}: small (124M), medium (350M), and large (760M), on FineWeb~\citep{penedo2024fineweb} at a $1\times$ Chinchilla compute budget ($\approx\!20$ tokens per parameter)~\citep{hoffmann2022training}, giving $2.49$B, $7.10$B, and $15.49$B training tokens over $4{,}750$, $13{,}550$, and $29{,}550$ optimizer steps, respectively, at a fixed batch of $0.5$M tokens (1k sequence length). All runs use bfloat16 mixed precision on $8{\times}$H100 (FSDP~\citep{zhao2023fsdp} at medium and large), with cosine decay to $10\%$ of the peak LR after $2\%$ warmup, gradient clipping at $1.0$, and $\varepsilon\!=\!10^{-8}$.

\noindent\textbf{Parameter groups.} Following the setup of~\citet{jordan2024muon, matrixwhitening2025}, only the matrix parameters (Q/K/V/O attention projections and MLP linears) are preconditioned by the optimizer under study, while embeddings, the lm-head, and 1D and norm parameters are routed to a fixed AdamW group with LR $10^{-2}$, $\beta\!=\!(0.95,0.95)$, no weight decay on embeddings, and weight decay $10^{-3}$ on the lm-head. Unless noted, all groups use $\beta\!=\!(0.95,0.95)$, and SOAP variants use precondition frequency $\mathtt{pf} = 10$ with the Purifying-Shampoo skip tolerance $\mathrm{tol} = 0.1$~\citep{2025purifying}.

\noindent\textbf{Hyperparameter search.} At small we run a $5{\times}5$ LR/WD grid (factor $10^{1/8}\!\approx\!1.33$ on LR as in~\citet{matrixwhitening2025}, and $10^{1/4}\!\approx\!1.78$ on WD) per optimizer at $\beta\!=\!(0.95,0.95)$, re-centering on grid edges, and re-validate the inner $3{\times}3$ neighborhood over three seeds. For medium and large, dense LR/WD grids are infeasible, so we transfer the small inner LR via a $\mu$P~\citep{yang2022tensor} ($\propto\!1/\text{width}$) anchor, sweep $5$ LR points (factor $10^{1/8}$) around it with WD held at the small inner optimum, refine the best point with a $3{\times}3$ LR/WD grid extended at grid edges, and report the selected cell as the per-size point estimate. Validation loss is reported as mean $\pm$ std over $11$ seeds at small and medium and $3$ seeds at large.

\noindent\textbf{Baselines.} The headline comparison spans AdamW~\citep{loshchilov2019adamw}, Muon~\citep{jordan2024muon}, AdaMuon~\citep{si2025adamuon}, SOAP~\citep{vyas2025soap} (the bilateral, dense-$v$ variant), and FOAM (ours) at all three scales. At GPT-2 small we additionally evaluate SPlus~\citep{frans2025splus}, Shampoo~\citep{gupta2018shampoo}, and the SOAP variants of our construction ablation (\S\ref{sec:exp:e3-stack}).

\subsection{Main Results}
\label{sec:exp:main}
Table~\ref{tab:main} reports the main pre-training results across three model scales. SOAP attains the lowest validation loss at every scale, with FOAM behind by $0.004$, $0.008$, and $0.005$ nats at small, medium, and large, each within about one seed std. FOAM is ahead of Muon, AdaMuon, and AdamW at every scale. The cost gap to SOAP \emph{widens at scale}. FOAM uses half of SOAP's matrix-parameter state at every size, and its measured peak memory at small size is $+7.4\%$ over AdamW against $+12.9\%$ for SOAP. On wall-clock time, FOAM needs fewer steps than AdamW to reach AdamW's final loss, which offsets its higher per-step cost: its time to that loss, $t_{\mathrm{A}}$, is within $7\%$ of AdamW's own at every scale and equal at large ($10.53$ vs.\ $10.55$\,h), whereas SOAP needs $1.35\times$ at small and $2.15\times$ at large, as its two eigendecompositions grow cubically with width.

\keyfinding{FOAM stays within $0.004$--$0.008$ nats of SOAP at every scale with half of SOAP's optimizer state, and reaches AdamW's final loss within $7\%$ of AdamW's own wall-clock time, while SOAP needs $2.15\times$ as long at 760M parameters.}

\subsection{Hyperparameter Robustness}
\label{sec:exp:hp}
\begin{figure}[t]
\centering
\includegraphics{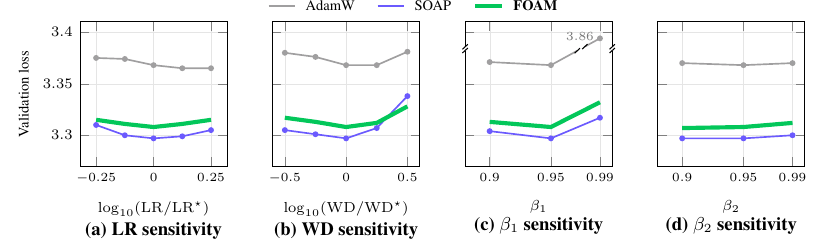}
\caption{\textbf{Hyperparameter robustness around each optimizer's best cell on GPT-2 small.} For AdamW, SOAP, and FOAM we sweep $5$ LR points (factor $10^{1/8}$), $5$ WD points (factor $10^{1/4}$) at the best LR, and $3$ values each of $\beta_1$ and $\beta_2$ at the best LR/WD. All other HPs are held fixed. Panel (c) breaks the $y$-axis at the top: AdamW's point at $\beta_1 = 0.99$ is $3.86$, far above the other values.}
\label{fig:hp-sensitivity}

\end{figure}

A practical optimizer must also tolerate imprecise tuning. Figure~\ref{fig:hp-sensitivity} sweeps LR, WD, and $\beta_1,\beta_2$ around each optimizer's inner best on GPT-2 small. FOAM's bowl shape and width track AdamW and SOAP across all four axes within $0.05$ nats. The one notable asymmetry is AdamW's catastrophic collapse to $3.86$ at $\beta_1\!=\!0.99$, where SOAP and FOAM degrade by only $0.02$ nats, so FOAM inherits SOAP's wider safe range on the momentum axis, as panel (c) shows.

\begin{table}[t]
\centering
\small
\caption{\textbf{FOAM-construction ablation on GPT-2 small (124M).} Starting from SOAP, input $R$ and output $L$ mark which Kronecker factor (with its rotation) is kept, and factored $v$ replaces the dense second moment by $r, c$. Each variant uses its own best LR/WD cell; mean $\pm$ std over 3 seeds of the ablation runs' final evaluation. state is the matrix-parameter optimizer state (GiB) and wall is the total training wall-clock on a single $8{\times}$H100 node.}
\label{tab:e3-stack}
\vspace{1mm}
\begin{tabular}{c c c c c c}
\toprule
 input $R$ & output $L$ & factored\,$v$
                 & val loss & state (GiB) & wall (h) \\
\midrule
$\checkmark$ & $\checkmark$ &              & $3.295{\scriptstyle\pm.011}$ & $2.92$ & $0.48$ \\
$\checkmark$ & $\checkmark$ & $\checkmark$ & $\bestm{3.293}{\scriptstyle\pm.008}$ & $2.59$ & $0.49$ \\
$\checkmark$ &              &              & $3.302{\scriptstyle\pm.008}$ & $1.78$ & $\bestm{0.44}$ \\
             & $\checkmark$ &              & $3.322{\scriptstyle\pm.010}$ & $1.78$ & $0.46$ \\
\oursrow
$\checkmark$ &              & $\checkmark$ & $3.304{\scriptstyle\pm.007}$ & $\bestm{1.46}$ & $0.46$ \\
\bottomrule
\end{tabular}%
\end{table}

\subsection{Analysis}
\label{sec:exp:analysis}
We first measure what each FOAM simplification contributes to quality and cost, then ask whether input-only preconditioning is structurally justified on actual GPT-2 layers, and finally whether the factored update stays close to the dense SOAP update in practice.

\noindent\textbf{FOAM-construction ablation.}
\label{sec:exp:e3-stack}
Table~\ref{tab:e3-stack} ablates, starting from SOAP, the two structural simplifications of FOAM on GPT-2 small, namely input-only preconditioning, which drops the output-side rotation $Q_L$, and the factored second moment, which replaces the dense $v$ with the row and column estimators $r, c$. Each variant is evaluated at its own best HP cell with $3$ seeds. Input-only preconditioning substantially reduces both optimizer state and wall-clock while keeping validation loss within $0.01$ nats of full SOAP, whereas output-only preconditioning, at the same state, loses about $0.02$ nats, in line with the input-output asymmetry analyzed below. Factorization further trades a small wall-clock increase for a meaningful reduction in optimizer state with no measurable loss in validation. Stacked, the two simplifications yield FOAM, which retains SOAP's validation while halving the matrix-parameter optimizer state. The same holds at medium and large, where input-only SOAP and FOAM differ by at most $0.01$ nats while the factored second moment saves $18\%$ of the state.

\begin{figure}[t]
\centering

\begin{minipage}[t]{0.49\linewidth}
\centering
\includegraphics{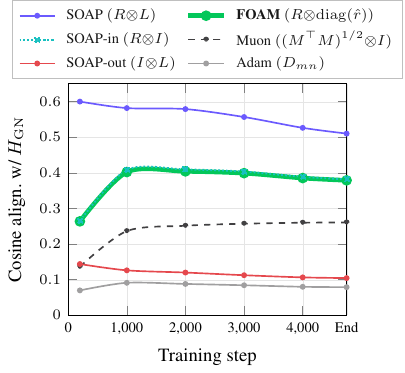}
\caption{\textbf{Layer-wise Gauss-Newton alignment.} For each of $72$ matrix parameters, we estimate $H_{\mathrm{GN}}\!=\!J_\ell^\top G_y J_\ell$ via $10$ to $30$ Hutchinson probes with the closed-form softmax-CE output Hessian, and compute the Frobenius cosine $\cos(P, H_{\mathrm{GN}})$ in sqrt-power form for six implicit preconditioners. Curves are means over the $72$ matrices along one SOAP run on GPT-2 small.}
\label{fig:fit-error}
\end{minipage}%
\hfill
\begin{minipage}[t]{0.49\linewidth}
\centering
\includegraphics{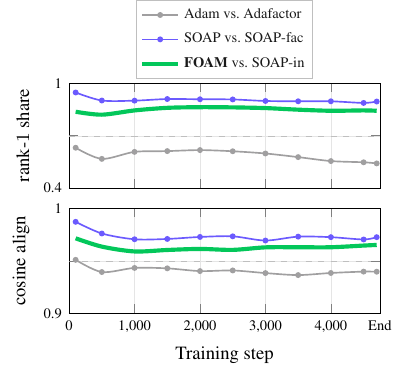}
\caption{\textbf{Validity of the factored second moment.} On the dense run of each regime, we form the rank-$1$ approximation $V_{\mathrm{fac}}\!=\!(V\mathbf{1})(\mathbf{1}^\top V)/(\mathbf{1}^\top V \mathbf{1})$ from the row and column sums of the second-moment estimate $V$, and report \textbf{(a)} the rank-$1$ share $\sigma_1(V)/\!\sum_i\!\sigma_i(V)$ and \textbf{(b)} the Frobenius cosine between $V_{\mathrm{fac}}$ and $V$, averaged across all matrix parameters.}
\label{fig:factored-validity}
\end{minipage}
\end{figure}

\noindent\textbf{Gauss-Newton fit.}
\label{sec:exp:e1}
Figure~\ref{fig:fit-error} compares six implicit preconditioners against the per-layer Gauss-Newton matrix $H_{\mathrm{GN}}$ on all $72$ matrix parameters of the $12$-layer GPT-2 small backbone. SOAP's full bilateral preconditioner is the closest Kronecker approximant of $H_{\mathrm{GN}}$, about $6\times$ above Adam-diag and $2\times$ above Muon. The asymmetry between input and output rotations is sharp and persistent. Dropping the output-side rotation costs only a quarter of the alignment, so input-only keeps $75\%$ of SOAP-full at $0.38$, whereas dropping the input-side rotation costs $80\%$ and leaves output-only at $0.10$. FOAM, which is input-only and factored, tracks SOAP-input-only within $1$--$2\%$ at every checkpoint, indicating that factoring the eigenvalue tensor preserves nearly all of input-only's curvature alignment. This asymmetry, in which the input-feature axis carries the dominant curvature structure, is the empirical justification for the input-only recipe used in the main results. It also holds per module, since on the exact layer-wise Gauss-Newton matrix input-only exceeds output-only in all six module types at both 124M and 350M, by $1.5\times$ to $11.5\times$ (Appendix~\ref{sec:appx:asym}).

\noindent\textbf{Factored update validity.}
\label{sec:exp:e2}
Figure~\ref{fig:factored-validity} reports two diagnostics that explain why the factored variants of Table~\ref{tab:e3-stack} preserve quality. The rank-$1$ share in panel (a) is the key diagnostic. In Adam's raw $g^2$ basis, $V$ is only $0.54$--$0.63$ rank-$1$ separable and drifts down during training, while in SOAP's eigenbasis the rotated $V$ stays near $0.90$ for the bilateral rotation and $0.84$ for the input-only rotation of FOAM. The cosine in panel (b) follows accordingly, at about $0.97$, $0.96$, and $0.94$ in the bilateral, input-only, and raw bases. Consequently, factoring the SOAP-rotated $V$ loses essentially nothing, costing less than $0.005$ nats in validation, whereas Adafactor-style factoring~\citep{shazeer2018adafactor} of Adam's raw $V$ is far less accurate. With the $H_{\mathrm{GN}}$ alignment of Figure~\ref{fig:fit-error}, this completes FOAM's rationale, which rotates into the input-side eigenbasis, where $V$ becomes nearly rank-$1$ separable, and then factors.

\keyfinding{The input side carries the curvature structure: input-only preconditioning keeps about $75\%$ of SOAP's Gauss-Newton alignment and output-only about $20\%$. After the input-side rotation the second moment is nearly rank-one separable, so at 124M factoring it costs only about $0.002$ nats while saving $18\%$ of the optimizer state.}

\subsection{Beyond the Main Setting}
\label{sec:exp:beyond}
\emph{Scale.} At GPT-2 XL, with $1.56$B parameters and $31$B tokens, FOAM and SOAP are within $0.001$ nats over three seeds and both ahead of Muon and AdamW, which are the other two optimizers run at this scale (Appendix~\ref{sec:appx:xl}).
\emph{Modality.} On ViT-Tiny trained on Tiny-ImageNet, the matrix-preconditioned optimizers beat AdamW by $6.5$ to $8.5$ points of top-1 accuracy, with SOAP first and FOAM second, on par with Muon and AdaMuon (Appendix~\ref{sec:appx:vit}).
\emph{Batch size.} At a fixed token budget, growing the batch from $0.5$M to $2$M tokens per step raises AdamW's loss from $3.370$ to $3.956$ but FOAM's only from $3.309$ to $3.364$. FOAM also stays ahead of Muon in final loss at every batch size, and at $2$M tokens per step only FOAM reaches AdamW's $0.5$M-batch quality. Larger batches thus stay usable with FOAM, which at $2$M tokens per step reaches AdamW's $0.5$M-batch quality in about a third of its own $0.5$M-batch wall-clock time (Appendix~\ref{sec:appx:batch}).

\keyfinding{The pre-training ordering carries over to a 1.56B model and to ViT classification, and FOAM keeps its quality at batch sizes where AdamW degrades sharply.}

\section{Discussion and Conclusion}
\label{sec:disc}

\noindent\textbf{Limitations and future work.} Our experiments stop at GPT-2 scale, up to 1.56B parameters, so whether FOAM's advantages hold for frontier-scale pre-training remains the main open question. Our implementation does not yet shard $R$ and $Q_R$ across devices, which matters at the $\geq 7$B scale, although the layer-wise distributed optimizer of~\citet{khona2026beyond} suggests that this is an engineering rather than a fundamental obstacle. The factored update is least accurate at the key projection, whose gradients pass through RoPE and softmax (Appendix~\ref{sec:appx:kfac}). Finally, the embeddings and the lm-head stay on AdamW because their vocabulary dimension rules out a dense factor on that side, although a FOAM-style factor on their small hidden dimension would be affordable. Extending FOAM to these layers, and attention-aware factorizations for the key projection, are natural next steps.

\noindent\textbf{When to use FOAM.} FOAM fits when SOAP-class quality is needed but SOAP's memory or wall-clock is binding, since it stays within $0.004$--$0.008$ nats of SOAP at half its optimizer state and, at 760M, about half its wall-clock. Against Muon and AdaMuon, quality slightly favors FOAM at every scale and speed is near parity, while AdaMuon keeps less state. Muon and AdaMuon thus remain preferable when optimizer state is the hardest constraint. Against AdamW, FOAM, like other matrix-preconditioned methods, is most valuable at large effective batch, where AdamW degrades (Appendix~\ref{sec:appx:batch}) and, at fixed $\mathtt{pf}$, the optimizer's compute share shrinks. FOAM is least attractive when memory is dominated by modules with a large input dimension, such as the MLP down-projection, whose $n^2$ factor carries most of FOAM's residual state over AdamW.

\noindent\textbf{Conclusion.} We presented \emph{FOAM}, a scalable and practical SOAP variant built on two structural simplifications: input-only preconditioning, justified by the Kronecker structure of the Hessian and a near-diagonal output curvature, and a factored Adam-Moment update, exact under K-FAC, with Purifying Shampoo's skip rule for the amortized eigenbasis update. Where memory-efficient variants mainly shrink AdamW's memory, FOAM addresses the complementary problem and brings SOAP-level quality close to Adam-level cost. FOAM matches strong Muon and SOAP baselines on LLM pre-training with modern architectures while using half of SOAP's optimizer state and Muon-level wall-clock time, from \texttt{124M} to \texttt{760M}, with supporting evidence at 1.56B and on ViT.

{
\small
\bibliographystyle{plainnat}
\bibliography{references}

\begin{thebibliography}{53}
\providecommand{\natexlab}[1]{#1}
\providecommand{\url}[1]{\texttt{#1}}
\expandafter\ifx\csname urlstyle\endcsname\relax
  \providecommand{\doi}[1]{doi: #1}\else
  \providecommand{\doi}{doi: \begingroup \urlstyle{rm}\Url}\fi

\bibitem[Agarwal et~al.(2024)Agarwal, Vieillard, Zhou, Stanczyk, Ramos~Garea, Geist, and Bachem]{agarwal2024gkd}
Rishabh Agarwal, Nino Vieillard, Yongchao Zhou, Piotr Stanczyk, Sabela Ramos~Garea, Matthieu Geist, and Olivier Bachem.
\newblock On-policy distillation of language models: Learning from self-generated mistakes.
\newblock In \emph{International Conference on Learning Representations (ICLR)}, 2024.
\newblock arXiv:2306.13649.

\bibitem[Amsel et~al.(2026)Amsel, Persson, Musco, and Gower]{amsel2026polar}
Noah Amsel, David Persson, Christopher Musco, and Robert~M. Gower.
\newblock The {Polar Express}: Optimal matrix sign methods and their application to the {Muon} algorithm.
\newblock In \emph{International Conference on Learning Representations (ICLR)}, 2026.
\newblock arXiv:2505.16932.

\bibitem[Anil et~al.(2020)Anil, Gupta, Koren, Regan, and Singer]{anil2020scalable}
Rohan Anil, Vineet Gupta, Tomer Koren, Kevin Regan, and Yoram Singer.
\newblock Scalable second order optimization for deep learning.
\newblock \emph{arXiv preprint arXiv:2002.09018}, 2020.

\bibitem[Bernstein et~al.(2018)Bernstein, Wang, Azizzadenesheli, and Anandkumar]{bernstein2018signum}
Jeremy Bernstein, Yu-Xiang Wang, Kamyar Azizzadenesheli, and Animashree Anandkumar.
\newblock {signSGD}: Compressed optimisation for non-convex problems.
\newblock In \emph{International Conference on Machine Learning (ICML)}, pages 560--569, 2018.

\bibitem[Collobert(2004)]{collobert2004large}
Ronan Collobert.
\newblock \emph{Large Scale Machine Learning}.
\newblock PhD thesis, Universit{\'e} Paris VI, Paris, France, 2004.

\bibitem[Dahl et~al.(2023)Dahl, Schneider, Nado, Agarwal, Sastry, Hennig, Medapati, Eschenhagen, Kasimbeg, Suo, Bae, Gilmer, Peirson, Khan, Anil, Rabbat, Krishnan, Snider, Amid, Chen, Maddison, Vasudev, Badura, Garg, and Mattson]{dahl2023benchmarking}
George~E. Dahl, Frank Schneider, Zachary Nado, Naman Agarwal, Chandramouli~Shama Sastry, Philipp Hennig, Sourabh Medapati, Runa Eschenhagen, Priya Kasimbeg, Daniel Suo, Juhan Bae, Justin Gilmer, Abel~L. Peirson, Bilal Khan, Rohan Anil, Mike Rabbat, Shankar Krishnan, Daniel Snider, Ehsan Amid, Kongtao Chen, Chris~J. Maddison, Rakshith Vasudev, Michal Badura, Ankush Garg, and Peter Mattson.
\newblock Benchmarking neural network training algorithms.
\newblock \emph{arXiv preprint arXiv:2306.07179}, 2023.

\bibitem[Dosovitskiy et~al.(2021)Dosovitskiy, Beyer, Kolesnikov, Weissenborn, Zhai, Unterthiner, Dehghani, Minderer, Heigold, Gelly, Uszkoreit, and Houlsby]{dosovitskiy2021vit}
Alexey Dosovitskiy, Lucas Beyer, Alexander Kolesnikov, Dirk Weissenborn, Xiaohua Zhai, Thomas Unterthiner, Mostafa Dehghani, Matthias Minderer, Georg Heigold, Sylvain Gelly, Jakob Uszkoreit, and Neil Houlsby.
\newblock An image is worth 16x16 words: Transformers for image recognition at scale.
\newblock In \emph{International Conference on Learning Representations (ICLR)}, 2021.
\newblock arXiv:2010.11929.

\bibitem[Duchi et~al.(2011)Duchi, Hazan, and Singer]{duchi2011adaptive}
John Duchi, Elad Hazan, and Yoram Singer.
\newblock Adaptive subgradient methods for online learning and stochastic optimization.
\newblock \emph{Journal of Machine Learning Research}, 12\penalty0 (61):\penalty0 2121--2159, 2011.

\bibitem[Duvvuri et~al.(2024)Duvvuri, Devvrit, Anil, Hsieh, and Dhillon]{duvvuri2024caspr}
Sai~Surya Duvvuri, Fnu Devvrit, Rohan Anil, Cho-Jui Hsieh, and Inderjit~S. Dhillon.
\newblock Combining axes preconditioners through {Kronecker} approximation for deep learning.
\newblock In \emph{International Conference on Learning Representations (ICLR)}, 2024.

\bibitem[Eschenhagen et~al.(2023)Eschenhagen, Immer, Turner, Schneider, and Hennig]{eschenhagen2023kfac}
Runa Eschenhagen, Alexander Immer, Richard~E. Turner, Frank Schneider, and Philipp Hennig.
\newblock {Kronecker}-factored approximate curvature for modern neural network architectures.
\newblock In \emph{Advances in Neural Information Processing Systems (NeurIPS)}, volume~36, 2023.
\newblock arXiv:2311.00636.

\bibitem[Eschenhagen et~al.(2025)Eschenhagen, Defazio, Lee, Turner, and Shi]{2025purifying}
Runa Eschenhagen, Aaron Defazio, Tsung-Hsien Lee, Richard~E. Turner, and Hao-Jun~Michael Shi.
\newblock Purifying {Shampoo}: Investigating {Shampoo}'s heuristics by decomposing its preconditioner.
\newblock In \emph{Advances in Neural Information Processing Systems (NeurIPS)}, 2025.
\newblock arXiv:2506.03595.

\bibitem[Frans et~al.(2025{\natexlab{a}})Frans, Abbeel, and Levine]{matrixwhitening2025}
Kevin Frans, Pieter Abbeel, and Sergey Levine.
\newblock What really matters in matrix-whitening optimizers?
\newblock \emph{arXiv preprint arXiv:2510.25000}, 2025{\natexlab{a}}.

\bibitem[Frans et~al.(2025{\natexlab{b}})Frans, Levine, and Abbeel]{frans2025splus}
Kevin Frans, Sergey Levine, and Pieter Abbeel.
\newblock A stable whitening optimizer for efficient neural network training.
\newblock In \emph{Advances in Neural Information Processing Systems (NeurIPS)}, volume~38, 2025{\natexlab{b}}.
\newblock arXiv:2506.07254.

\bibitem[George et~al.(2018)George, Laurent, Bouthillier, Ballas, and Vincent]{george2018ekfac}
Thomas George, C{\'e}sar Laurent, Xavier Bouthillier, Nicolas Ballas, and Pascal Vincent.
\newblock Fast approximate natural gradient descent in a {Kronecker}-factored eigenbasis.
\newblock In \emph{Advances in Neural Information Processing Systems (NeurIPS)}, volume~31, 2018.
\newblock arXiv:1806.03884.

\bibitem[Gong et~al.(2026)Gong, Scetbon, Ma, and Meeds]{gong2025alice}
Wenbo Gong, Meyer Scetbon, Chao Ma, and Edward Meeds.
\newblock Towards efficient optimizer design for {LLM} via structured {Fisher} approximation with a low-rank extension.
\newblock In \emph{International Conference on Learning Representations (ICLR)}, 2026.
\newblock arXiv:2502.07752.

\bibitem[Gupta et~al.(2018)Gupta, Koren, and Singer]{gupta2018shampoo}
Vineet Gupta, Tomer Koren, and Yoram Singer.
\newblock {Shampoo}: Preconditioned stochastic tensor optimization.
\newblock In \emph{International Conference on Machine Learning (ICML)}, pages 1842--1850, 2018.
\newblock arXiv:1802.09568.

\bibitem[Hendrycks and Gimpel(2016)]{hendrycks2016gelu}
Dan Hendrycks and Kevin Gimpel.
\newblock Gaussian error linear units ({GELUs}).
\newblock \emph{arXiv preprint arXiv:1606.08415}, 2016.

\bibitem[Hendrycks et~al.(2021)Hendrycks, Burns, Kadavath, Arora, Basart, Tang, Song, and Steinhardt]{hendrycks2021math}
Dan Hendrycks, Collin Burns, Saurav Kadavath, Akul Arora, Steven Basart, Eric Tang, Dawn Song, and Jacob Steinhardt.
\newblock Measuring mathematical problem solving with the {MATH} dataset.
\newblock In \emph{Advances in Neural Information Processing Systems (NeurIPS), Datasets and Benchmarks Track}, volume~1, 2021.
\newblock arXiv:2103.03874.

\bibitem[Hoffmann et~al.(2022)Hoffmann, Borgeaud, Mensch, Buchatskaya, Cai, Rutherford, de~Las~Casas, Hendricks, Welbl, Clark, Hennigan, Noland, Millican, van~den Driessche, Damoc, Guy, Osindero, Simonyan, Elsen, Vinyals, Rae, and Sifre]{hoffmann2022training}
Jordan Hoffmann, Sebastian Borgeaud, Arthur Mensch, Elena Buchatskaya, Trevor Cai, Eliza Rutherford, Diego de~Las~Casas, Lisa~Anne Hendricks, Johannes Welbl, Aidan Clark, Thomas Hennigan, Eric Noland, Katherine Millican, George van~den Driessche, Bogdan Damoc, Aurelia Guy, Simon Osindero, Karen Simonyan, Erich Elsen, Oriol Vinyals, Jack~W. Rae, and Laurent Sifre.
\newblock An empirical analysis of compute-optimal large language model training.
\newblock In \emph{Advances in Neural Information Processing Systems (NeurIPS)}, volume~35, 2022.
\newblock arXiv:2203.15556 (arXiv version titled ``Training Compute-Optimal Large Language Models'').

\bibitem[Jordan et~al.(2024)Jordan, Jin, Boza, You, Cesista, Newhouse, and Bernstein]{jordan2024muon}
Keller Jordan, Yuchen Jin, Vlado Boza, Jiacheng You, Franz Cesista, Laker Newhouse, and Jeremy Bernstein.
\newblock Muon: An optimizer for hidden layers in neural networks.
\newblock \url{https://kellerjordan.github.io/posts/muon/}, 2024.
\newblock Blog post, December 2024. Accessed: 2026-05-07.

\bibitem[Kasimbeg et~al.(2025)Kasimbeg, Schneider, Eschenhagen, Bae, Sastry, Saroufim, Feng, Wright, Yang, Nado, Medapati, Hennig, Rabbat, and Dahl]{kasimbeg2025algoperf}
Priya Kasimbeg, Frank Schneider, Runa Eschenhagen, Juhan Bae, Chandramouli~Shama Sastry, Mark Saroufim, Boyuan Feng, Less Wright, Edward~Z. Yang, Zachary Nado, Sourabh Medapati, Philipp Hennig, Michael Rabbat, and George~E. Dahl.
\newblock Accelerating neural network training: An analysis of the {AlgoPerf} competition.
\newblock In \emph{International Conference on Learning Representations (ICLR)}, 2025.
\newblock arXiv:2502.15015.

\bibitem[Khona et~al.(2026)Khona, Vavre, Wang, Fu, Wu, Chrzanowski, Catanzaro, Mudigere, Pool, Lightstone, Shoeybi, Patwary, Tajbakhsh, and Blankevoort]{khona2026beyond}
Mikail Khona, Aditya Vavre, Boxiang Wang, Deyu Fu, Hao Wu, Mike Chrzanowski, Bryan Catanzaro, Dheevatsa Mudigere, Jeff Pool, Michael Lightstone, Mohammad Shoeybi, Mostofa Patwary, Nima Tajbakhsh, and Tijmen Blankevoort.
\newblock {SOAP}, {Muon}, and beyond: Pushing {LLM} pretraining scales.
\newblock \emph{arXiv preprint arXiv:2607.20548}, 2026.

\bibitem[Kingma and Ba(2015)]{kingma2015adam}
Diederik~P. Kingma and Jimmy Ba.
\newblock {Adam}: A method for stochastic optimization.
\newblock In \emph{International Conference on Learning Representations (ICLR)}, 2015.
\newblock arXiv:1412.6980.

\bibitem[Kunstner et~al.(2019)Kunstner, Hennig, and Balles]{kunstner2019empirical}
Frederik Kunstner, Philipp Hennig, and Lukas Balles.
\newblock Limitations of the empirical {Fisher} approximation for natural gradient descent.
\newblock In \emph{Advances in Neural Information Processing Systems (NeurIPS)}, volume~32, 2019.
\newblock arXiv:1905.12558.

\bibitem[Kunstner et~al.(2024)Kunstner, Milligan, Yadav, Schmidt, and Bietti]{kunstner2024whyadam}
Frederik Kunstner, Alan Milligan, Robin Yadav, Mark Schmidt, and Alberto Bietti.
\newblock Heavy-tailed class imbalance and why {Adam} outperforms gradient descent on language models.
\newblock In \emph{Advances in Neural Information Processing Systems (NeurIPS)}, volume~37, pages 30106--30148, 2024.
\newblock arXiv:2402.19449.

\bibitem[Le and Yang(2015)]{le2015tiny}
Ya~Le and Xuan Yang.
\newblock Tiny {ImageNet} visual recognition challenge.
\newblock CS231N course project report, Stanford University, 2015.
\newblock URL \url{http://cs231n.stanford.edu/reports/2015/pdfs/yle_project.pdf}.

\bibitem[Li et~al.(2024)Li, Beeching, Tunstall, Lipkin, Soletskyi, Huang, Rasul, Yu, Jiang, Shen, Qin, Dong, Zhou, Fleureau, Lample, and Polu]{li2024numinamath}
Jia Li, Edward Beeching, Lewis Tunstall, Ben Lipkin, Roman Soletskyi, Shengyi~Costa Huang, Kashif Rasul, Longhui Yu, Albert Jiang, Ziju Shen, Zihan Qin, Bin Dong, Li~Zhou, Yann Fleureau, Guillaume Lample, and Stanislas Polu.
\newblock {NuminaMath}.
\newblock \url{https://github.com/project-numina/aimo-progress-prize/blob/main/report/numina_dataset.pdf}, 2024.
\newblock Hugging Face dataset: \url{https://huggingface.co/datasets/AI-MO/NuminaMath-CoT}.

\bibitem[Li et~al.(2026)Li, Liu, Liang, Chen, and Zhao]{li2026normuon}
Zichong Li, Liming Liu, Chen Liang, Weizhu Chen, and Tuo Zhao.
\newblock {NorMuon}: Making {Muon} more efficient and scalable.
\newblock In \emph{International Conference on Machine Learning (ICML)}, pages 68514--68533, 2026.
\newblock arXiv:2510.05491.

\bibitem[Lightman et~al.(2024)Lightman, Kosaraju, Burda, Edwards, Baker, Lee, Leike, Schulman, Sutskever, and Cobbe]{lightman2024verify}
Hunter Lightman, Vineet Kosaraju, Yuri Burda, Harrison Edwards, Bowen Baker, Teddy Lee, Jan Leike, John Schulman, Ilya Sutskever, and Karl Cobbe.
\newblock Let's verify step by step.
\newblock In \emph{International Conference on Learning Representations (ICLR)}, 2024.
\newblock arXiv:2305.20050.

\bibitem[Lin et~al.(2026)Lin, Lowe, Dangel, Eschenhagen, Xu, and Grosse]{lin2025kl}
Wu~Lin, Scott~C. Lowe, Felix Dangel, Runa Eschenhagen, Zikun Xu, and Roger~B. Grosse.
\newblock Understanding and improving {Shampoo} and {SOAP} via {Kullback}-{Leibler} minimization.
\newblock In \emph{International Conference on Learning Representations (ICLR)}, 2026.
\newblock arXiv:2509.03378.

\bibitem[Liu et~al.(2026)Liu, Zhang, Wang, Zhao, Su, Yang, and Zhang]{liu2026muon2}
Ziyue Liu, Ruijie Zhang, Zhengyang Wang, Yequan Zhao, Yupeng Su, Zi~Yang, and Zheng Zhang.
\newblock {Muon}$^2$: Boosting {Muon} via adaptive second-moment preconditioning.
\newblock \emph{arXiv preprint arXiv:2604.09967}, 2026.

\bibitem[Loshchilov and Hutter(2019)]{loshchilov2019adamw}
Ilya Loshchilov and Frank Hutter.
\newblock Decoupled weight decay regularization.
\newblock In \emph{International Conference on Learning Representations (ICLR)}, 2019.
\newblock arXiv:1711.05101.

\bibitem[Lu et~al.(2025)Lu, Wang, and Liu]{lu2025soap}
Yanqing Lu, Letao Wang, and Jinbo Liu.
\newblock Understanding {SOAP} from the perspective of gradient whitening.
\newblock \emph{arXiv preprint arXiv:2509.22938}, 2025.

\bibitem[Ma et~al.(2025)Ma, Gong, Scetbon, and Meeds]{ma2024swan}
Chao Ma, Wenbo Gong, Meyer Scetbon, and Edward Meeds.
\newblock {SWAN}: {SGD} with normalization and whitening enables stateless {LLM} training.
\newblock In \emph{International Conference on Machine Learning (ICML)}, pages 41907--41942, 2025.
\newblock arXiv:2412.13148.

\bibitem[Martens and Grosse(2015)]{martens2015kfac}
James Martens and Roger Grosse.
\newblock Optimizing neural networks with {Kronecker}-factored approximate curvature.
\newblock In \emph{International Conference on Machine Learning (ICML)}, pages 2408--2417, 2015.

\bibitem[Morwani et~al.(2025)Morwani, Shapira, Vyas, Malach, Kakade, and Janson]{morwani2024shampoo}
Depen Morwani, Itai Shapira, Nikhil Vyas, Eran Malach, Sham Kakade, and Lucas Janson.
\newblock A new perspective on {Shampoo}'s preconditioner.
\newblock In \emph{International Conference on Learning Representations (ICLR)}, 2025.
\newblock arXiv:2406.17748.

\bibitem[Penedo et~al.(2024)Penedo, Kydl{\'\i}{\v{c}}ek, Ben~Allal, Lozhkov, Mitchell, Raffel, Von~Werra, and Wolf]{penedo2024fineweb}
Guilherme Penedo, Hynek Kydl{\'\i}{\v{c}}ek, Loubna Ben~Allal, Anton Lozhkov, Margaret Mitchell, Colin Raffel, Leandro Von~Werra, and Thomas Wolf.
\newblock The {FineWeb} datasets: Decanting the web for the finest text data at scale.
\newblock In \emph{Advances in Neural Information Processing Systems (NeurIPS), Datasets and Benchmarks Track}, volume~37, pages 30811--30849, 2024.
\newblock arXiv:2406.17557.

\bibitem[Radford et~al.(2019)Radford, Wu, Child, Luan, Amodei, and Sutskever]{radford2019language}
Alec Radford, Jeffrey Wu, Rewon Child, David Luan, Dario Amodei, and Ilya Sutskever.
\newblock Language models are unsupervised multitask learners.
\newblock \emph{OpenAI blog}, 1\penalty0 (8):\penalty0 9, 2019.

\bibitem[Shazeer and Stern(2018)]{shazeer2018adafactor}
Noam Shazeer and Mitchell Stern.
\newblock {Adafactor}: Adaptive learning rates with sublinear memory cost.
\newblock In \emph{International Conference on Machine Learning (ICML)}, pages 4596--4604, 2018.
\newblock arXiv:1804.04235.

\bibitem[Shi et~al.(2023)Shi, Lee, Iwasaki, Gallego-Posada, Li, Rangadurai, Mudigere, and Rabbat]{shi2023distributed}
Hao-Jun~Michael Shi, Tsung-Hsien Lee, Shintaro Iwasaki, Jose Gallego-Posada, Zhijing Li, Kaushik Rangadurai, Dheevatsa Mudigere, and Michael Rabbat.
\newblock A distributed data-parallel {PyTorch} implementation of the distributed {Shampoo} optimizer for training neural networks at-scale.
\newblock \emph{arXiv preprint arXiv:2309.06497}, 2023.

\bibitem[Si et~al.(2025)Si, Zhang, and Shen]{si2025adamuon}
Chongjie Si, Debing Zhang, and Wei Shen.
\newblock {AdaMuon}: Adaptive {Muon} optimizer.
\newblock \emph{arXiv preprint arXiv:2507.11005}, 2025.

\bibitem[Su et~al.(2024)Su, Ahmed, Lu, Pan, Bo, and Liu]{su2024roformer}
Jianlin Su, Murtadha Ahmed, Yu~Lu, Shengfeng Pan, Wen Bo, and Yunfeng Liu.
\newblock {RoFormer}: Enhanced transformer with rotary position embedding.
\newblock \emph{Neurocomputing}, 568:\penalty0 127063, 2024.
\newblock arXiv:2104.09864.

\bibitem[Touvron et~al.(2021)Touvron, Cord, Douze, Massa, Sablayrolles, and J{\'e}gou]{touvron2021deit}
Hugo Touvron, Matthieu Cord, Matthijs Douze, Francisco Massa, Alexandre Sablayrolles, and Herv{\'e} J{\'e}gou.
\newblock Training data-efficient image transformers \& distillation through attention.
\newblock In \emph{International Conference on Machine Learning}, pages 10347--10357. PMLR, 2021.

\bibitem[Vyas et~al.(2025)Vyas, Morwani, Zhao, Shapira, Brandfonbrener, Janson, and Kakade]{vyas2025soap}
Nikhil Vyas, Depen Morwani, Rosie Zhao, Itai Shapira, David Brandfonbrener, Lucas Janson, and Sham Kakade.
\newblock {SOAP}: Improving and stabilizing {Shampoo} using {Adam} for language modeling.
\newblock In \emph{International Conference on Learning Representations (ICLR)}, 2025.
\newblock arXiv:2409.11321.

\bibitem[Yang et~al.(2025)Yang, Li, Yang, Zhang, Hui, Zheng, Yu, Gao, Huang, Lv, Zheng, Liu, Zhou, Huang, Hu, Ge, Wei, Lin, Tang, Yang, Tu, Zhang, Yang, Yang, Zhou, Zhou, Lin, Dang, Bao, Yang, Yu, Deng, Li, Xue, Li, Zhang, Wang, Zhu, Men, Gao, Liu, Luo, Li, Tang, Yin, Ren, Wang, Zhang, Ren, Fan, Su, Zhang, Zhang, Wan, Liu, Wang, Cui, Zhang, Zhou, and Qiu]{yang2025qwen3}
An~Yang, Anfeng Li, Baosong Yang, Beichen Zhang, Binyuan Hui, Bo~Zheng, Bowen Yu, Chang Gao, Chengen Huang, Chenxu Lv, Chujie Zheng, Dayiheng Liu, Fan Zhou, Fei Huang, Feng Hu, Hao Ge, Haoran Wei, Huan Lin, Jialong Tang, Jian Yang, Jianhong Tu, Jianwei Zhang, Jianxin Yang, Jiaxi Yang, Jing Zhou, Jingren Zhou, Junyang Lin, Kai Dang, Keqin Bao, Kexin Yang, Le~Yu, Lianghao Deng, Mei Li, Mingfeng Xue, Mingze Li, Pei Zhang, Peng Wang, Qin Zhu, Rui Men, Ruize Gao, Shixuan Liu, Shuang Luo, Tianhao Li, Tianyi Tang, Wenbiao Yin, Xingzhang Ren, Xinyu Wang, Xinyu Zhang, Xuancheng Ren, Yang Fan, Yang Su, Yichang Zhang, Yinger Zhang, Yu~Wan, Yuqiong Liu, Zekun Wang, Zeyu Cui, Zhenru Zhang, Zhipeng Zhou, and Zihan Qiu.
\newblock {Qwen3} technical report.
\newblock \emph{arXiv preprint arXiv:2505.09388}, 2025.

\bibitem[Yang et~al.(2021)Yang, Hu, Babuschkin, Sidor, Liu, Farhi, Ryder, Pachocki, Chen, and Gao]{yang2022tensor}
Ge~Yang, Edward Hu, Igor Babuschkin, Szymon Sidor, Xiaodong Liu, David Farhi, Nick Ryder, Jakub Pachocki, Weizhu Chen, and Jianfeng Gao.
\newblock Tuning large neural networks via zero-shot hyperparameter transfer.
\newblock In \emph{Advances in Neural Information Processing Systems (NeurIPS)}, volume~34, pages 17084--17097, 2021.
\newblock arXiv:2203.03466 (extended arXiv version titled ``Tensor Programs V: ...'').

\bibitem[Zhang and Sennrich(2019)]{zhang2019rmsnorm}
Biao Zhang and Rico Sennrich.
\newblock Root mean square layer normalization.
\newblock In \emph{Advances in Neural Information Processing Systems (NeurIPS)}, volume~32, 2019.
\newblock arXiv:1910.07467.

\bibitem[Zhang et~al.(2026)Zhang, Liu, and Schaeffer]{zhang2026namo}
Minxin Zhang, Yuxuan Liu, and Hayden Schaeffer.
\newblock {Adam} improves {Muon}: Adaptive moment estimation with orthogonalized momentum.
\newblock \emph{arXiv preprint arXiv:2602.17080}, 2026.

\bibitem[Zhang et~al.(2025{\natexlab{a}})Zhang, Xu, Luo, and Wang]{zhang2025depthwise}
Tianxiao Zhang, Wenju Xu, Bo~Luo, and Guanghui Wang.
\newblock Depth-wise convolutions in vision transformers for efficient training on small datasets.
\newblock \emph{Neurocomputing}, 617:\penalty0 128998, 2025{\natexlab{a}}.
\newblock arXiv:2407.19394.

\bibitem[Zhang et~al.(2025{\natexlab{b}})Zhang, Chen, Li, Ding, Wu, Kingma, Ye, Luo, and Sun]{zhang2025adammini}
Yushun Zhang, Congliang Chen, Ziniu Li, Tian Ding, Chenwei Wu, Diederik~P. Kingma, Yinyu Ye, Zhi-Quan Luo, and Ruoyu Sun.
\newblock {Adam-mini}: Use fewer learning rates to gain more.
\newblock In \emph{International Conference on Learning Representations (ICLR)}, 2025{\natexlab{b}}.
\newblock arXiv:2406.16793.

\bibitem[Zhao et~al.(2024)Zhao, Zhang, Chen, Wang, Anandkumar, and Tian]{zhao2024galore}
Jiawei Zhao, Zhenyu Zhang, Beidi Chen, Zhangyang Wang, Anima Anandkumar, and Yuandong Tian.
\newblock {GaLore}: Memory-efficient {LLM} training by gradient low-rank projection.
\newblock In \emph{International Conference on Machine Learning (ICML)}, pages 61121--61143, 2024.

\bibitem[Zhao et~al.(2023)Zhao, Gu, Varma, Luo, Huang, Xu, Wright, Shojanazeri, Ott, Shleifer, Desmaison, Balioglu, Damania, Nguyen, Chauhan, Hao, Mathews, and Li]{zhao2023fsdp}
Yanli Zhao, Andrew Gu, Rohan Varma, Liang Luo, Chien-Chin Huang, Min Xu, Less Wright, Hamid Shojanazeri, Myle Ott, Sam Shleifer, Alban Desmaison, Can Balioglu, Pritam Damania, Bernard Nguyen, Geeta Chauhan, Yuchen Hao, Ajit Mathews, and Shen Li.
\newblock {PyTorch FSDP}: Experiences on scaling fully sharded data parallel.
\newblock \emph{Proceedings of the VLDB Endowment}, 16\penalty0 (12):\penalty0 3848--3860, 2023.
\newblock arXiv:2304.11277.

\bibitem[Zhu et~al.(2025)Zhu, Zhang, Cong, Liu, Park, Chandra, Long, Pan, Wang, and Lee]{zhu2025apollo}
Hanqing Zhu, Zhenyu Zhang, Wenyan Cong, Xi~Liu, Sem Park, Vikas Chandra, Bo~Long, David~Z. Pan, Zhangyang Wang, and Jinwon Lee.
\newblock {APOLLO}: {SGD}-like memory, {AdamW}-level performance.
\newblock In \emph{Proceedings of Machine Learning and Systems (MLSys)}, volume~7, 2025.
\newblock arXiv:2412.05270.

\end{thebibliography}
}

\clearpage
\appendix
\setcounter{table}{0}\renewcommand{\thetable}{\Alph{table}}\renewcommand{\theHtable}{app.\Alph{table}}
\setcounter{figure}{0}\renewcommand{\thefigure}{\Alph{figure}}\renewcommand{\theHfigure}{app.\Alph{figure}}
\section{Selected Hyperparameters}
\label{sec:appx:protocol}

Table~\ref{tab:hp-cells} lists the matrix-group LR and WD that each optimizer selects in the search of \S\ref{sec:exp:setup}.

\begin{table}[ht]
\centering
\small
\caption{\textbf{Selected matrix-group LR / WD} behind Table~\ref{tab:main}, for each optimizer and scale; all use $\beta = (0.95, 0.95)$ and $\varepsilon = 10^{-8}$, and the tail group is fixed as in \S\ref{sec:exp:setup}.}
\label{tab:hp-cells}
\begin{tabular}{l ccc}
\toprule
Optimizer & small (124M) & medium (350M) & large (760M) \\
\midrule
AdamW   & $1.9\mathrm{e}{-3}$ / $0.316$   & $1.06\mathrm{e}{-3}$ / $0.316$  & $8.5\mathrm{e}{-4}$ / $0.178$ \\
Muon    & $1.92\mathrm{e}{-2}$ / $0.0562$ & $1.09\mathrm{e}{-2}$ / $0.0316$ & $8.76\mathrm{e}{-3}$ / $0.0178$ \\
AdaMuon & $4.41\mathrm{e}{-4}$ / $3.16$   & $3.33\mathrm{e}{-4}$ / $1.0$    & $2.0\mathrm{e}{-4}$ / $1.0$ \\
SPlus   & $0.333$ / $0.00562$             & --                              & -- \\
Shampoo & $1.4\mathrm{e}{-2}$ / $0.1$     & --                              & -- \\
SOAP    & $1.86\mathrm{e}{-3}$ / $0.562$  & $3.32\mathrm{e}{-3}$ / $0.1$    & $1.99\mathrm{e}{-3}$ / $0.1$ \\
\oursrow
\textbf{FOAM (ours)} & $2.5\mathrm{e}{-3}$ / $0.316$ & $1.41\mathrm{e}{-3}$ / $0.178$ & $1.51\mathrm{e}{-3}$ / $0.1$ \\
\bottomrule
\end{tabular}
\end{table}

\section{Shampoo on GPT-2 Small}
\label{sec:appx:shampoo}

We run Shampoo~\citep{gupta2018shampoo} at small only, as the Distributed Shampoo preconditioner with exponent $-1/4$ per factor and no grafting, and otherwise with the recipe and tuning of the other optimizers (Table~\ref{tab:hp-cells}). Shampoo lands $0.07$ nats behind FOAM and even slightly behind AdamW (Table~\ref{tab:shampoo}), whereas SOAP, which runs Adam in the same Kronecker eigenbasis, closes this gap.

\begin{table}[ht]
\centering
\small
\caption{\textbf{Shampoo on GPT-2 small (124M).} Validation loss at each optimizer's selected cell.}
\label{tab:shampoo}
\begin{tabular}{l c}
\toprule
Optimizer & val \\
\midrule
AdamW   & $3.370{\scriptstyle\pm.009}$ \\
Shampoo & $3.379{\scriptstyle\pm.006}$ \\
SOAP    & $\bestm{3.305}{\scriptstyle\pm.009}$ \\
\oursrow
\textbf{FOAM (ours)} & $3.309{\scriptstyle\pm.008}$ \\
\bottomrule
\end{tabular}
\end{table}

\section{Validity of the K-FAC Assumption}
\label{sec:appx:kfac}

The factored update is exact under the K-FAC factorization (Lemma~\ref{lem:rotcov}); the diagonal output side matters only for matching SOAP's two-sided rotation (Appendix~\ref{sec:appx:asym}). For each instrumented matrix we measure on checkpoints
(i) the \emph{fidelity} $\rho(W) = \| \Lambda - r c^\top \|_F / \| \Lambda \|_F$, the relative Frobenius error of the row-column factorization of the rotated second moment $\Lambda$ ($\rho = 0$ under K-FAC), and
(ii) the \emph{sensitivity} $\Delta(W) = \cos(P_{\mathrm{dense}} V, H_{\mathrm{GN}} V) - \cos(P_{\mathrm{fac}} V, H_{\mathrm{GN}} V)$ over random probes $V$, where $H_{\mathrm{GN}}$ is the layer's exact Gauss-Newton matrix (labels sampled from the model) and $P_{\mathrm{dense}}$, $P_{\mathrm{fac}}$ are FOAM's preconditioner with the dense and the factored second moment, so that $\Delta$ isolates what the violation costs the update. The $18$ instrumented matrices are the six module types at three depths, at 124M and 350M. Table~\ref{tab:kfac-layer} reports every matrix at step $2{,}000$.

\paragraph{Findings.}
The violation is moderate and scale-stable: $\rho$ averages $0.162$ at 124M and $0.147$ at 350M, and the row-column marginals nearly reach the best rank-one fit (rank-one energy $0.975$). The attention key projection is the outlier ($\rho$ up to $0.79$ in middle and late layers; its gradients pass through RoPE and softmax), while $W_v$ and $W_o$ are almost exact. The consequence for the update is small: $\Delta$ averages $+0.005$ and $+0.004$, and stays below $+0.05$ even for $W_k$, consistent with the $0.002$-nat end-to-end difference between input-only SOAP and FOAM (Table~\ref{tab:e3-stack}). Across six training stages at 124M (steps $200$--$4{,}749$), the $18$-matrix mean of $\rho$ stays within $0.16$--$0.19$ and that of $\Delta$ within $+0.001$--$+0.005$, and at the three stages measured at 350M within $0.15$--$0.17$ and $+0.003$--$+0.004$, so no stage of training makes the factored update meaningfully worse.

\begin{table}[ht]
\centering
\footnotesize
\caption{\textbf{Per-layer K-FAC violation $\rho$ and its cost $\Delta$} (shown as $\rho$\,/\,$\Delta$) at step $2{,}000$ for every instrumented matrix: blocks $0$, $6$, $11$ of 124M and $0$, $12$, $23$ of 350M.}
\label{tab:kfac-layer}
\tabcolsep=0.35em
\begin{tabular}{l ccc ccc}
\toprule
& \multicolumn{3}{c}{124M} & \multicolumn{3}{c}{350M} \\
\cmidrule(lr){2-4} \cmidrule(lr){5-7}
module & early & mid & late & early & mid & late \\
\midrule
attention $W_q$ & $.209$\,/\,$+.001$ & $.059$\,/\,$+.000$ & $.067$\,/\,$+.001$ & $.350$\,/\,$-.001$ & $.070$\,/\,$-.001$ & $.062$\,/\,$-.001$ \\
attention $W_k$ & $.372$\,/\,$+.005$ & $.785$\,/\,$+.044$ & $.769$\,/\,$+.046$ & $.352$\,/\,$+.004$ & $.706$\,/\,$+.036$ & $.508$\,/\,$+.030$ \\
attention $W_v$ & $.017$\,/\,$+.000$ & $.026$\,/\,$-.000$ & $.011$\,/\,$-.000$ & $.027$\,/\,$+.001$ & $.028$\,/\,$-.000$ & $.012$\,/\,$-.000$ \\
attention $W_o$ & $.013$\,/\,$-.000$ & $.027$\,/\,$-.001$ & $.023$\,/\,$-.001$ & $.014$\,/\,$-.000$ & $.029$\,/\,$-.001$ & $.020$\,/\,$-.001$ \\
MLP up         & $.035$\,/\,$-.000$ & $.077$\,/\,$-.001$ & $.122$\,/\,$+.002$ & $.039$\,/\,$-.000$ & $.087$\,/\,$-.000$ & $.060$\,/\,$+.004$ \\
MLP down       & $.050$\,/\,$-.000$ & $.145$\,/\,$-.002$ & $.112$\,/\,$+.000$ & $.084$\,/\,$-.000$ & $.154$\,/\,$-.001$ & $.047$\,/\,$+.001$ \\
\midrule
mean over $18$ & \multicolumn{3}{c}{$.162$\,/\,$+.005$} & \multicolumn{3}{c}{$.147$\,/\,$+.004$} \\
\bottomrule
\end{tabular}
\end{table}

\section{Per-module Input/Output Asymmetry}
\label{sec:appx:asym}

Table~\ref{tab:gn-permodule} breaks the Gauss-Newton alignment down by module type, measured with the exact layer-wise Gauss-Newton matrix of Appendix~\ref{sec:appx:kfac} at step $2{,}000$ (Figure~\ref{fig:fit-error} uses a Hutchinson estimate over all $72$ matrices of a separate run). Removing the output-side rotation from the two-sided preconditioner costs $0.160$ of alignment on average, whereas removing the input-side rotation costs $0.375$; averaged over depth, input-only exceeds output-only in every module type at both scales ($1.5\times$ to $10.9\times$ at 124M and $1.9\times$ to $11.5\times$ at 350M), although a few individual layers reverse the order. Output-only preconditioning ($0.132$) is barely above diagonal Adam ($0.090$). The two factors differ in conditioning: after normalizing each by its own diagonal, the ratio of the largest to the $20$th eigenvalue is $5.0$ on the output side against $67.9$ on the input side ($7.4$ against $213$ on the EMA factors SOAP maintains), so on the output side the coordinate axes already capture most of the structure.

\begin{table}[ht]
\centering
\small
\caption{\textbf{Per-module input/output asymmetry at 124M.} Gauss-Newton alignment per module (mean over early, mid, and late depth) for output-only, input-only, and two-sided rotations; diagonal Adam scores $0.090$ on the same measure, and the 350M values give the same ordering.}
\label{tab:gn-permodule}
\begin{tabular}{l ccc c}
\toprule
module & output only & input only & both & input / output \\
\midrule
attention $W_q$ & $0.176$ & $0.322$ & $0.499$ & $1.8\times$ \\
attention $W_k$ & $0.118$ & $0.178$ & $0.314$ & $1.5\times$ \\
attention $W_v$ & $0.209$ & $0.453$ & $0.614$ & $2.2\times$ \\
attention $W_o$ & $0.045$ & $0.493$ & $0.629$ & $\bestm{10.9\times}$ \\
MLP up   & $0.152$ & $0.332$ & $0.522$ & $2.2\times$ \\
MLP down & $0.091$ & $0.300$ & $0.466$ & $3.3\times$ \\
\midrule
mean over $18$ matrices & $0.132$ & $0.347$ & $0.507$ & $2.6\times$ \\
\bottomrule
\end{tabular}
\end{table}

\section{GPT-2 XL (1.56B)}
\label{sec:appx:xl}

We test whether the ordering of the three smaller sizes holds at GPT-2 XL with $1.56$B parameters (Table~\ref{tab:xl}). SOAP and FOAM are tied (SOAP $-$ FOAM $= -0.0008 \pm 0.0002$, paired over seeds), and both are $0.012$--$0.013$ nats ahead of Muon and $0.04$ ahead of AdamW.

The model has $48$ layers and width $1600$ and otherwise follows \S\ref{sec:exp:setup}. It is trained for $59{,}450$ steps of $0.5$M tokens ($31$B tokens, $1\times$ Chinchilla) on the FineWeb $100$B-token sample with $2\%$ warmup and cosine decay to $10\%$, and evaluated on a fixed $20$M-token validation set. Each optimizer is tuned on $11$ to $14$ LR and WD cells with seed $42$, centered by the rule $\mathrm{LR} \cdot \mathrm{WD} \cdot T \approx \text{const}$ fitted on the smaller sizes, and its best cell is re-run on seeds $1339$ and $2024$. SOAP and FOAM use $\mathtt{pf} = 10$.

\begin{table}[ht]
\centering
\small
\caption{\textbf{GPT-2 XL (1.56B, $31$B tokens).} Validation loss on a fixed $20$M-token set, mean $\pm$ std over seeds $\{42, 1339, 2024\}$ at each optimizer's coarse-grid optimum.}
\label{tab:xl}
\begin{tabular}{l c c c}
\toprule
Optimizer & LR / WD & val \\
\midrule
AdamW & $2.4\mathrm{e}{-4}$ / $0.423$  & $2.6365{\scriptstyle\pm.0002}$ \\
Muon  & $6.78\mathrm{e}{-3}$ / $0.0089$ & $2.6081{\scriptstyle\pm.0005}$ \\
SOAP  & $1.17\mathrm{e}{-3}$ / $0.0529$ & $\bestm{2.5951}{\scriptstyle\pm.0007}$ \\
\oursrow
\textbf{FOAM (ours)} & $1.48\mathrm{e}{-3}$ / $0.043$ & $2.5959{\scriptstyle\pm.0003}$ \\
\bottomrule
\end{tabular}
\end{table}

\section{ViT-Tiny on Tiny-ImageNet}
\label{sec:appx:vit}

We test whether the comparison transfers across modality, task, and loss by training ViT-Tiny~\citep{dosovitskiy2021vit, touvron2021deit} for Tiny-ImageNet~\citep{le2015tiny} classification (Table~\ref{tab:vit}). The matrix-preconditioned optimizers beat AdamW ($58.6$) by $6.5$ to $8.5$ points of top-1 accuracy, SOAP leads ($67.1$), and FOAM is second ($65.4$), tied with Muon and AdaMuon (both $65.1$) within noise.

We follow the plain ViT-Tiny recipe of~\citet{zhang2025depthwise}, with $300$ epochs, a cosine schedule with $20$-epoch warmup, and its augmentation and regularization, but train at the native image size $64$ with patch size $4$ and batch $256$. Only the attention and MLP weight matrices use the optimizer under study, and all other parameters run AdamW with LR $5{\cdot}10^{-4}$ and WD $0.05$. Each optimizer's matrix LR and WD are tuned on a geometric grid with ratios $10^{1/4}$ and $10^{1/2}$ on one seed.

\begin{table}[ht]
\centering
\small
\caption{\textbf{ViT-Tiny on Tiny-ImageNet.} Top-1 is the best evaluation over $300$ epochs (final-epoch values are $0.1$--$0.4$ points lower, with SOAP first, FOAM second, and AdamW last); single seed. AdamW's matrix group uses $\beta = (0.9, 0.999)$, the others $(0.95, 0.95)$.}
\label{tab:vit}
\begin{tabular}{l c c}
\toprule
Optimizer & matrix LR / WD & top-1 (\%) \\
\midrule
AdamW   & $1.0\mathrm{e}{-3}$ / $0.158$  & $58.6$ \\
Muon    & $1.0\mathrm{e}{-2}$ / $0.0158$ & $65.1$ \\
AdaMuon & $1.0\mathrm{e}{-3}$ / $0.158$  & $65.1$ \\
SOAP    & $5.6\mathrm{e}{-3}$ / $0.05$   & $\bestm{67.1}$ \\
\oursrow
\textbf{FOAM (ours)} & $1.8\mathrm{e}{-3}$ / $0.158$ & $65.4$ \\
\bottomrule
\end{tabular}
\end{table}

\section{Batch-size Scaling}
\label{sec:appx:batch}

A fixed token budget can be spent at different batch sizes, and under data-parallel weak scaling each batch doubling roughly halves the wall-clock time, so we test whether that speed is usable (Table~\ref{tab:batch}). AdamW degrades sharply ($3.370 \to 3.465 \to 3.956$) and never recovers its $0.5$M-batch quality at larger batches, whose wall-clock saving thus comes with a worse model. FOAM holds ($3.309 \to 3.315 \to 3.364$) and stays ahead of Muon at every batch size, turning the same scaling into about a $3\times$ reduction in the time to AdamW's quality ($24.9 \to 13.8 \to 8.2$ minutes).

GPT-2 small follows the recipe of \S\ref{sec:exp:setup} with a fixed $2.49$B-token budget, i.e., $4{,}750$ steps at $0.5$M tokens, and each optimizer's LR is re-tuned at $0.5$M, $1$M, and $2$M tokens per step.

\begin{table}[ht]
\centering
\small
\caption{\textbf{Batch-size scaling on GPT-2 small} (under the fixed $2.49$B-token budget, LR re-tuned at every batch size). \emph{wall} $\rightarrow 3.370$ is the wall-clock time to reach AdamW's own $0.5$M-batch quality under data-parallel weak scaling (total wall $\propto$ steps); N/R (not reached) marks runs that do not reach it, and the $1$M and $2$M rows are from a single seed each.}
\label{tab:batch}
\begin{tabular}{l ccc cc}
\toprule
& \multicolumn{3}{c}{final val loss} & \multicolumn{2}{c}{wall $\rightarrow 3.370$} \\
\cmidrule(lr){2-4} \cmidrule(lr){5-6}
tokens/step (steps) & AdamW & Muon & FOAM & AdamW & FOAM \\
\midrule
$0.5$M ($4{,}750$) & $3.370$ & $3.313$ & $\bestm{3.309}$ & $23.8$ min & $24.9$ min \\
$1$M ($2{,}375$)   & $3.465$ & $3.323$ & $\bestm{3.315}$ & N/R & $13.8$ min \\
$2$M ($1{,}188$)   & $3.956$ & $3.384$ & $\bestm{3.364}$ & N/R & $8.2$ min \\
\bottomrule
\end{tabular}
\end{table}
\section{LLM Post-training}
\label{sec:appx:posttrain}

We test whether FOAM can replace AdamW in two post-training recipes on Qwen3~\citep{yang2025qwen3}, changing only the optimizer (Table~\ref{tab:posttrain}). FOAM is best in SFT and tied with AdamW in OPD, while Muon is $0.013$ below both in OPD. The SFT differences among the three optimizers are within $0.002$ NLL, so FOAM serves as a drop-in replacement for AdamW in both recipes.

Only the 2D weight matrices of the transformer blocks use the optimizer under study, and the rest run AdamW. One LR, shared by all groups, is swept per optimizer. FOAM uses $\mathtt{pf} = 10$ and $\mathrm{tol} = 0$, and Muon scales its update by $0.2\sqrt{\max(m, n)}$. All runs use no weight decay, gradient clipping at $1.0$, a cosine schedule to $10\%$, bfloat16 with fp32 master weights, and a single seed.
\emph{Supervised fine-tuning (SFT).} Qwen3-1.7B-Base is fine-tuned for one epoch on $100$K NuminaMath-CoT~\citep{li2024numinamath} examples ($782$ steps of batch $128$, up to $4{,}096$ tokens, $\beta = (0.9, 0.95)$) and evaluated by the NLL on $1{,}000$ held-out examples.
\emph{On-policy distillation (OPD)~\citep{agarwal2024gkd}.} Qwen3-1.7B is distilled from Qwen3-4B-Instruct-2507 in non-thinking mode with a per-token reverse-KL loss on its own rollouts for open reasoning prompts ($100$ steps of $256$ prompts, up to $8{,}192$ tokens, $\beta = (0.9, 0.999)$) and scored by average accuracy on MATH500~\citep{hendrycks2021math, lightman2024verify}, AIME24, AIME25, and AMC23.

\begin{table}[ht]
\centering
\small
\caption{\textbf{Post-training of Qwen3-1.7B with each optimizer at its best LR} (in parentheses). SFT reports held-out NLL (lower is better) and OPD average math accuracy (higher is better), single seed.}
\label{tab:posttrain}
\tabcolsep=0.4em
\begin{tabular}{l c ccc}
\toprule
Recipe & base & AdamW & Muon & FOAM \\
\midrule
SFT & $0.625$ & $0.4677$ ($3\mathrm{e}{-5}$) & $0.4668$ ($3\mathrm{e}{-5}$) & $\bestm{0.4657}$ ($3\mathrm{e}{-5}$) \\
OPD & $0.338$ & $\bestm{0.4719}$ ($1\mathrm{e}{-5}$) & $0.4591$ ($1\mathrm{e}{-5}$) & $\bestm{0.4719}$ ($1\mathrm{e}{-5}$) \\
\bottomrule
\end{tabular}
\end{table}

\end{document}